\documentclass[pdflatex,sn-mathphys-num]{sn-jnl}% Math and Physical Sciences Numbered Reference Style 
\usepackage{graphicx}%
\usepackage{multirow}%
\usepackage{amsmath,amssymb,amsfonts}%
\usepackage{amsthm}%
\usepackage{mathrsfs}%
\usepackage[title]{appendix}%
\usepackage{xcolor}%
\usepackage{textcomp}%
\usepackage{manyfoot}%
\usepackage{booktabs}%
\usepackage{algorithm}%
\usepackage{algorithmicx}%
\usepackage{algpseudocode}%
\usepackage{listings}%
\usepackage{amssymb}
\usepackage{pifont}
\usepackage{xcolor}
\theoremstyle{thmstyleone}%
\theoremstyle{thmstyletwo}%

\theoremstyle{thmstylethree}%

\begin{document}

\title[Article Title]{L-XAIDS: A LIME-based eXplainable AI framework for Intrusion Detection Systems}

%%=============================================================%%
%% GivenName	-> \fnm{Joergen W.}
%% Particle	-> \spfx{van der} -> surname prefix
%% FamilyName	-> \sur{Ploeg}
%% Suffix	-> \sfx{IV}
%% \author*[1,2]{\fnm{Joergen W.} \spfx{van der} \sur{Ploeg} 
%%  \sfx{IV}}\email{iauthor@gmail.com}
%%=============================================================%%

\author*[1]{\fnm{Aoun E} \sur{Muhammad}}\email{amo874@uregina.ca}

\author[1]{\fnm{Kin-Choong} \sur{Yow}}\email{kin.choong.yow@uregina.ca}

\author[2]{\fnm{Nebojša } \sur{Bačanin-Džakula}}\email{nbacanin@singidunum.ac.rs}
\author[3]{\fnm{Muhammad Attique} \sur{Khan}}\email{attique.khan@ieee.org}

\affil*[1]{\orgdiv{Faculty of Engineering and Applied Science}, \orgname{University of Regina, }, \orgaddress{\street{3737 Wascana Pkway}, \city{Regina}, \postcode{S4S 0A2}, \state{Saskatchewan}, \country{Canada}}}

\affil[2]{\orgdiv{Head of Applied Artificial Intelligence}, \orgname{Singidunum University}, \orgaddress{\street{Danijelova 32}, \city{Belgrade}, \postcode{160622}, \state{Beograd}, \country{Serbia}}}

\affil[3]{\orgdiv{Department of AI, College of Computer Engineering and Science}, \orgname{Prince Mohammad bin Fahd University}, \orgaddress{\street{617, Al Jawharah}, \city{Khobar}, \postcode{34754}, \state{Dhahran}, \country{Saudi Arabia}}}

%%==================================%%
%% Sample for unstructured abstract %%
%%==================================%%

\abstract{Recent developments in Artificial Intelligence (AI) and their applications in critical industries such as healthcare, fin-tech and cybersecurity have led to a surge in research in explainability in AI. Innovative research methods are being explored to extract meaningful insight from blackbox AI systems to make the decision-making technology transparent and interpretable. Explainability becomes all the more critical when AI is used in decision making in domains like fintech, healthcare and safety critical systems such as cybersecurity and autonomous vehicles. However, there is still ambiguity lingering on the reliable evaluations for the users and nature of transparency in the explanations provided for the decisions made by black-boxed AI. To solve the blackbox nature of Machine Learning based Intrusion Detection Systems, a framework is proposed in this paper to give an explanation for IDSs decision making. This framework uses Local Interpretable Model-Agnostic Explanations (LIME) coupled with Explain Like I'm five (ELI5) and Decision Tree algorithms to provide local and global explanations and improve the interpretation of IDSs. The local explanations provide the justification for the decision made on a specific input. Whereas, the global explanations provides the list of significant features and their relationship with attack traffic. In addition, this framework brings transparency in the field of ML driven IDS that might be highly significant for wide scale adoption of eXplainable AI in cyber-critical systems. Our framework is able to achieve 85 percent accuracy in classifying attack behaviour on UNSW-NB15 dataset, while at the same time displaying the feature significance ranking of the top 10 features used in the classification.}


\maketitle

\section{Introduction}
The most crucial attribute for data-driven mission critical systems is their ability to explain their decisions to overseers. This also includes plausible explanation of their inner functionality and the metrics being used to make a decision or a prediction. This transparency for data-driven machine learning models becomes even more critical in the case of mission critical systems especially those relevant to cybersecurity. Decision or predictions made by the model should be easily comprehensible by the human, hence, simplifying the working of the model for the overseer to understand. 

While various explainability approaches have been explored for IDSs, existing methods often fail to provide actionable insights for cybersecurity professionals. Many black-box IDS models lack transparency, making it difficult to validate decisions, troubleshoot false positives, or ensure regulatory compliance. Furthermore, most explainability frameworks focus on either local or global explanations but fail to integrate both perspectives for a holistic understanding of IDS behavior. Our framework, L-XAIDS, addresses these gaps by incorporating LIME and ELI5, allowing cybersecurity teams to interpret both instance-specific and dataset-wide attack detection rationales.

The higher level of explainability helps in building trust on the systems working and suggestions made by them. When these models are fed with zero-day exploits or data they are not trained on, there might be a possibility that these systems might falsely flag the data packets as malicious or misclassify these instances. The magnitude and impact of misclassification may have serious repercussions as classifying an attack as normal traffic may eventually lead to a breach in the network. The process of flagging these misclassifications and trying to develop an understanding for the reasons this misclassification happened is the first step of improving these systems. For this process to be carried out successfully, clear and elaborate explanations are needed that may lead to bring improvements to the detection systems as well as assist in preventing future breaches. These explanations will assist the security professionals to study, intervene and overturn the decision being made by analyzing the weighted attributes for local and global explanations.

As cyber networks hold an ever-increasing amount of valuable user data, they have become a more valuable target for malicious attacks  \cite{C1}. It has high significance to critical services  where cyber-attackers attempt to compromise the security principles of confidentiality , integrity, and availability. In response, network intrusion detection systems (IDS) or systems that monitor and detect cyber-attack patterns over networking environments based on machine learning algorithms/models have been developed to detect suspicious network activity where They monitor network traffic for suspicious activities and issue alerts in case of detected attack types. 

Although ML algorithms and models can identify malicious activity or botnets by learning the cyber network inputs linked with Normal or Attack behaviors, current research indicates that these IDSs are generally regarded as "black boxes"\cite{C2}\cite{C3}. This means that cybersecurity service users are unable to comprehend the reasoning behind the system's attack prediction or classification, which is crucial for optimal initial cybersecurity evaluation and information assurance planning, as well as resource allocation for networks. The demand for robust cybersecurity practices and systems has risen with the widespread use of smart devices in various applications and settings. 

Extraordinary advancement in the domains of parallel hardware\cite{R1} \cite{R2} \cite{R3} and scalable systems \cite{R4}\cite{R5}\cite{R6} has resulted in real-time machine learning and AI algorithms \cite{R7} \cite{R8} \cite{R9}\cite{R10}\cite{R11}\cite{R12}. These advancements have resulted in such mature AI systems that can surpass human capabilities when it comes to train and process large scale critical operations data. These next generation AI algorithms are contributing massively in fields like cybersecurity, web search, smart cities, healthcare systems, IoT enabled environments and fast paced commerce/trading. All these aforementioned fields lie in the domain of mission critical systems. All these fields pose real time challenges for AI models such as:

\begin{itemize}
\item Decision making in real-time environments, 
\item Robust and securely designed models for predictions to filter out the adversarial inputs or maliciously intended data to be fed to the systems, 
\item Adding explainability to the decisions being made for the overseer to understand the working of the model and interpret the decision transparently.
\end{itemize}

A User’s understanding of a system can be defined as his ability to interpret the workings of a system and how it behaves in an explicit situation and why it suggests what it suggests and based on what metrics. The persuasive reasoning of an explanation is needed to convince the overseer to follow the given the novel circumstances.
Task performance, on other hand, is a combination of system’s precision, true/false positives/negatives, the explanation it provided for decision making processes and user’s understanding of the system. Combining all these concepts together builds up  trust on an AI system, that is an important topic in XAI research. If a user is well aware of system’s processes and algorithm metrics, it can be deduced that this understanding will help in building trust in the predictions being made by the system and will also improve task performance. The variation in the nature of explanations provided and the trust instill in those explanation will affect the users’ perception of the performance of the system resulting in judging them as well trusted or mis-trusted systems that could also result in affecting the task performance. \cite{R13}.

Our framework, L-XAIDS, directly addresses these concerns by integrating local (LIME) and global (ELI5) explainability methods into IDSs. While prior research has attempted to enhance IDS interpretability, existing approaches often focus on a single explainability technique, making it difficult for security analysts to gain a holistic view of model decisions. L-XAIDS fills this gap by providing a dual-explanation framework that enables both feature-level attribution (local) and dataset-wide analysis (global). By leveraging decision trees alongside these methods, our approach ensures that interpretability does not compromise predictive accuracy, which remains a limitation in some earlier works.

\subsection{Contributions}\label{subsec1}

\quad The major contribution of this research is to provide explanations to the decisions made by IDS on both local and global level. This explainability, in turn, enables overseers to understand the working and predictions made by the IDSs and bring transparecny in the decision making process. This framework also assists the cybersecurity professionals to learn the characterstics to a specific type of cyberattack.  This research aims to bring transparency to the IDS, as it presents the explainability in the form and combination of LIME-ELI5 technique to explain the decisions made by IDSs and bring transparency to the predictions.

This research work presents the following contributions:
\begin {enumerate}
\item Combined Approach: L-XAIDS leverages both LIME (for local explanations) and ELI5 (for global explanations) techniques. This combination provides a more comprehensive view of decision-making factors compared to frameworks solely focused on one approach. 
\begin{itemize}
\item Local explanations: Detail the relevance of each feature value to a specific packet's classification (positive or negative).
\item Global explanations: Highlight important dataset features that drive overall decision-making.
\end{itemize}
\item Enhanced Transparency and Performance: L-XAIDS not only improves transparency for human oversight but also aims to enhance overall IDS performance through better understanding of decision-making processes. This can lead to improved tasks like event management, logging, and identifying false positives/negatives.
\begin{itemize}
\item Enables security professionals to understand IDS predictions and decision-making processes.
\item Helps identify lesser-known attacks by analyzing Indicators of Compromise (IoCs).
\item Improves tasks like event management, real-time monitoring, and logging for auditing.
\item Assists in understanding false positives/negatives and system behavior in specific scenarios.
\end{itemize}

\end {enumerate}

These local and global level explanations contribute in improving task performances such as event management, real-time monitoring, logging and tracking the predictions for auditing and compliance purposes. These explanations will also contribute in identifying the relatively lesser-known attacks by analyzing the indicators of compromise (IoC) of certain malware types and their noteworthy attack vectors. Futhermore, these explanations can also assist to develop an understanding of a decision making system to identify the false/true negatives/positives predicted and the way it would behave in specific circumstances.

\section{Related Work}\label{sec2}

\quad The wide-scale adoption of automated systems for monitoring and control of critical infrastructure has made them vulnerable to cyber-attacks and opened up new attach surface to be exploited such as: intercepting, removing or replacing information without authorization, penetrating systems, creating backdoors and injecting viruses \cite{R14}\cite{R15}\cite{R16}\cite{R17}.
Deploying IDSs is one of the most effective mitigation techniques to detect and defend mission critical infrastructure against malicious attacks \cite{R18}. The two most essential preferences for an effective IDS are scalability and adaptability \cite{R19}. The sole purpose of zero-day attacks and exploits is to deceive these detection/prevention systems and successfully intrude the network. For this very reason, real-time machine learning driven cyber secure systems are greatly in demand nowadays \cite{R19}.
\subsection{Deep Learning driven black-boxed IDS algorithms}
\quad Some of the machine learning techniques like deep neural networks are being used in Intrusion detection and prevention systems as of now \cite{R19}\cite{R20}. The problem with these already in use algorithms is lack of transparency and their inherent black boxness. \cite{R20}\cite{R21}. Explainable AI in mission critical systems refers to an interface or level of explainability provided to the user that helps him understand the functioning of algorithm and the metrics that were considered while making a decision or a prediction \cite{R22}.  In recent years there has been a surge in research in the field of expalainable AI \cite{R22} and it has come to a point where governments are pushing for legislations for transparency in machine learning decision support systems \cite{R23}\cite{R24}. 
\subsection{IDS for smart environments}
\quad Protecting Internet of Things (IoT) networks has been a critical are of research for cybersecurity experts as IoT continues to be integrated in applications including modern healthcare systems, smart cities, home automation, and advanced manufacturing. Furthermore, developing machine learning based IDSs techniques and other statistical  feature learning algorithms has been researched and been put into application minimally. A study conducted by Nour Mustafa \cite{R25}developed an ensemble-based machine learning intrusion detection algorithm to protect the IoT network traffic based on statistical flow features. In his proposed system, statistical flow features were generated from initial analysis of network features. Thereafter, an AdaBoost ensemble learning method was applied to three machine learning algorithms including Naive Bayes (NB), decision tree,  and artificial neural network. The developed models were analyzed for performance to detect malicious activity rigorously, trained on the UNSW-NB15 and NIMS botnet datasets. The experimental results conveyed high performance of detection of normal and malicious activity. Overall, the proposed ensemble technique has a higher detection rate and a lower false positive rate when compared to three other traditional cybersecurity techniques. Kelton da Costa's survey reveals that numerous intrusion detection systems, based on machine learning, have been the subject of extensive research and development.\cite{R26}. 

The role of machine learning-based IDSs in Software-Defined Networking (SDN) environments has gained traction in recent years. A comprehensive review \cite{E4} outlines the recent advancements and challenges in deploying IDS solutions in SDN. While SDN-based IDSs benefit from adaptive detection mechanisms, their reliance on complex models necessitates the integration of explainability frameworks such as L-XAIDS to improve analyst trust and decision-making.

\subsection{Explainability and Interpretability for IDS systems}
\quad Recently, XAI had gained notable momentum, as lack of explainability has become an inherent problem of the latest ML techniques such as Deep Neural Networks (DNN) or ensembles. Hence, XAI, has become crucial for  deployments of ML models, where researchers keep model explainability, fairness, transparency and accountability at its core. More specifically, early investigation into developing Network Intrusion Detection System using an Explainable AI Frameworks has conducted by Shraddha Mane and Dattaraj Rao \cite{R27}. 
ML models tend to have higher accuracy as their complexity increases, but this often comes at the cost of reduced interpretability. In their study, the authors created a deep neural network for detecting network intrusions and proposed an explainable AI framework to maintain model transparency throughout the machine learning pipeline. To accomplish this, they utilized existing XAI algorithms such as SHAP, CEM (Contrastive Explanations Method) and LIME, which can provide explanations for individual predictions. They applied these approaches to the NSL-KDD dataset and successfully increased the model transparency, as evidenced by the results of their study.
Some of the IDS algorithms and solutions proposed in recent past used shallow machine learning algorithms to make predictions. On the other hand, algorithms proposed in  \cite{R28} \cite{R29} used Decision tree to make predictions. In \cite{R30}\cite{R31} SVM has been used to detect malicious algroithm. Whereas in  \cite{R32}\cite{R33}, KNN and Bayes Classifier methods have been applied respectively for malicious packet detection process.
In some other researches feature selection\cite{R34} and ensemble methods \cite{R35} have been used for intrusion detection process . 
In past few years, cybersecurity researchers have starting exploring DL techniques such  as deep neural network \cite{R36}, convolutional neural networks  \cite{R37}, recurrent neural network \cite{R38}\cite{R39}, variational autoencoder \cite{R40}\cite{R41}, etc for intrusion detection. Deep learning based IDSs systems demonstrate exceptional malicious packet detection accuracy and lower false flagging rate \cite{R42}.

For instance, deep learning has been applied to intrusion detection in in-vehicle networks, as demonstrated by \cite{E1}. This study highlights how deep learning models can effectively classify malicious and normal in-vehicle network traffic. However, despite its accuracy, deep learning models suffer from interpretability issues, reinforcing the need for explainability techniques such as those integrated into L-XAIDS

In \cite{R43} Besharati et. al proposed a host-based intrusion detection system for cloud environments, leveraging logistic regression. However, the lack of explainability method and anlaysis makes it difficult to understand and interpret the model's decision-making process.

In \cite{R44} Liu et. al proposed introduced a scalable method for host-based intrusion detection systems, employing Apache Spark alongside a range of machine learning techniques like decision trees, random forest, and support vector machines. However, the authors do not provide any explainability method or analysis in their paper. 

In \cite{R45}  an approach to intrusion detection using a Siamese network was proposed. The model learns similarity scores between a sequence of system calls and the previously learned normal behavior patterns. The research focuses only on local explainability utilizing a method called Grad-CAM, which highlights the most important system calls in the detection process.

In \cite{R46} an intrusion detection method using few-shot learning was proposed. The proposed method leverages transfer learning and uses only a few samples to train the model, thus overcoming the limitations of traditional machine learning-based intrusion detection methods that require large amounts of labeled data. 

In \cite{R47} authors proposed an intrusion detection method using supervised adversarial variational auto-encoder (SAVAE) with regularization.  The SAVAE model is used to learn the features of the normal network traffic data, and the anomalous traffic can be detected by measuring the difference between the learned features and the input data. The authors evaluated their proposed method on a benchmark dataset and compare it with several state-of-the-art intrusion detection methods.
In \cite{R48} authors proposed an adversarial attack on deep learning models used for network security to highlight the importance of robustness. The authors addressed the issue of robustness in deep-learning based network security models and proposed two methods: an adversarial training approach and a generative approach that generates adversarial samples to augment the training dataset. 

In \cite{R49} authors proposed a two-stage machine learning-based Wi-Fi network intrusion detection system using decision trees, random forest, and gradient boosting algorithms.  The paper does not provide a detailed explanation of the explainability methods used in the system, making it difficult to evaluate the system's explainability. The scope of explainability is not clearly defined, and it is not clear whether the system provides local or global explainability. 

A recent study \cite{E2} introduced a hybrid deep learning approach that leverages spatial-temporal features for intrusion detection. While this approach improves detection accuracy, it does not address the interpretability concerns inherent to deep learning models, making explainability techniques such as those in L-XAIDS crucial for real-world IDS applications.

In \cite{R50} authors evaluated the effectiveness of standard feature sets in ML-based network intrusion detection systems. They tested four different feature sets, including time-domain features, frequency-domain features, statistical features, and graph features, using random forest, k-nearest neighbors, and decision tree algorithms and LIME method for global explainability. They evaluate their system on UNSW-NB15 and CICIDS2017 datasets. The main advantage of this approach is that it provides good global explainability. However, the approach is limited to global explainability, and it is unclear how well it performs on other datasets. The paper evaluates the models' explainability using SHAP for global explanation but no discussion on local explanation
In \cite{R51} authors investigated the performance of machine learning and deep learning models for intrusion detection systems (IDSs) in terms of their accuracy and explainability.  The paper also investigates the impact of dataset characteristics on the performance of these models. The study found that deep learning models outperformed traditional machine learning models, and the performance of the models varied significantly depending on the dataset characteristics. 
In \cite{R52} authors proposed a novel approach to detect malware using a Graph Convolutional Network (GCN). The GCN is used to represent system-call dependency graphs and learn the hidden patterns in them. The proposed method is evaluated using the benchmark datasets from the malware detection community. One of the advantages of this approach is that it can detect zero-day malware, i.e., new malware that has never been seen before. However, the approach requires a large amount of training data and a significant computational cost. The interpretability of the model is limited, and it is challenging to extract meaningful explanations from the GCN model.
However, the inherent blackboxness of these deep learning methods pose an ongoing challenge for cybersecurity researchers. Security engineers' decision making process, nowadays, bank heavily on the predictions made by SIEM/IDS systems. Hence, making the working of system transparent becomes an issue of significant importance. Also, the complex nature of the prediction systems is a major penalty for the humans involved in the decision making/overseeing process as these blackbox models do not provide reasoning or justification for the decisions being made.  Hence, we believe it is crucial to provide the human-in-the-loop with some explanable information regarding the decisions being made by the IDS and provide cybersecurity engineers a better insight into the decision-making and prediction process. A few researches such as \cite{R53} \cite{R54} proposed adding explanations as a feature to IDSs, but lack the theoratical foundation provided by LIME.

Jiang et. al \cite{R55} proposed a network intrusion detection method that combines hybrid sampling and deep hierarchical networks. The explainability method used for this research is Grad-CAM , which provides a low level of global and local interpretation of the model's behavior. While the paper discusses the proposed system's overall performance, it does not delve into the explainability of the system in detail. The paper mentions that the proposed system's feature selection method reduces the feature dimensionality and improves the model's explainability. However, it does not explain the specific features that are selected or provide any further insight into the explainability of the feature selection process.  The deep hierarchical network architecture used in the system is briefly explained, but there is no mention of any specific explainability technique used to interpret the model's decision-making process. Therefore, the paper lacks detail on the system's explainability and the explanations provided for the intrusion detection. 

Wang et al. \cite{R56} focused on building an intrusion detection system that is explainable. The authors proposed an algorithm that combines decision tree and extreme gradient boosting to classify network traffic data and then uses SHAP (SHapley Additive exPlanations) to provide global explanations for the classification results. In terms of explainability, the paper provides detailed global and minimal local explanations for the classification results. The authors used SHAP to generate explanations for individual predictions as well as global explanations for the overall model. Does not comment or explain how local explanation is provided. Another limitation is that the paper does not address the trade-off between accuracy and explainability. However, the limitations of the paper suggest that further research is needed to fully evaluate the effectiveness of the proposed algorithm, and to determine the optimal method for providing local and global explanations in this use case.

Wali and Khan  \cite{R57} proposed an intrusion detection system based on Random Forests and SHAP for explainability. The paper only provides global explainability by utilizing SHAP values. The level of explainability can be considered as medium since the study provides feature importance and contribution towards the output for each feature. The main limitation of the study is that it does not discuss local explanations at all and only utilizes one explainability technique, which is SHAP values for providing global explanations. Although SHAP values are effective for explaining random forest-based models, they are not suitable for all models, and there is no one-size-fits-all explanation method. 

Tcydenova et al. \cite{R58} proposed an approach for detecting adversarial attacks in AI-based intrusion detection systems using explainable AI. The authors evaluated their approach on the NSL-KDD dataset using SVM, and LIME for explainability. The results show that their approach can detect adversarial attacks with medium level of explainability.  Local explanations are generated for individual predictions, while no global explanations is provided . The drawbacks and limitations of the paper include the small size of the dataset used for evaluation and the lack of comparison with other explainability techniques. 

Recent advancements in machine learning-based intrusion detection have focused on mitigating Advanced Persistent Threats (APTs)\cite{C7} and evolving cybersecurity strategies for AI-driven protection mechanisms\cite{C8}. Our work builds upon these contributions by integrating explainability techniques to enhance analyst decision-making in IDS environments.

Zebin et. al  \cite{R59} proposed an AI based explainable IDS for detecting DNS over HTTP (DoH) attacks. The authors used a DNS over HTTPS (DoH) attacks dataset extracted from CIRA-CIC-DoHBrw-2020 dataset  and applied Balanced Stacked random forest. SHAP was used for global explainability and the advantage of this research is the development of a new intrusion detection system that is explainable and can detect DoH attacks.  The system employs an ensemble model consisting of several machine learning models such as Random Forest, Gradient Boosting Machine, Logistic Regression, and Support Vector Machine, and uses SHAP values to provide global interpretability. However, the local explanations are not catered in this research.

Andresini et. al  \cite{R60} propose a neural attention-based intrusion detection system that can provide both local and global explainability. The authors used the neural attention algorithm for classification. They used integrated gradient for explainability with high level of explainability, used a multi-output approach and attention maps to provide explanations for its decisions. The authors claim that the attention mechanism used in their model provides better explainability compared to other approaches such as saliency maps and feature importance scores.  The attention mechanism used in the model provides global explanations with minimal localized explanations for its decisions. The proposed model requires a large amount of labeled training data, which can be difficult to obtain in practice and do not cater for potential biases of the attention mechanism used in the model.

Mowla et al.  \cite{R61} proposed a dynamic voting-based explainable intrusion detection system for in-vehicle network traffic. They used ensemble learning with a voting algorithm to develop the system and employed feature and model exploration for local explainability. The model's explainability was limited to the local scope, and the level of explainability was low. One of the major drawback is that the model's accuracy decreases when dealing with a large amount of data.

Mahbooba et al.  \cite{R62} explores the performance of machine learning and deep learning models for intrusion detection systems (IDSs). The authors compared the performance of various machine learning and deep learning models on different datasets. The paper also investigates the impact of dataset characteristics on the performance of these models.  The paper provides a comprehensive study on the performance analysis of machine learning and deep learning models for intrusion detection systems. The authors compare the performance of different models using five different datasets, which adds to the reliability of their findings. However, neither the explainability techniques for the evaluated method has been discussed nor any insights into the interpretability of the models or their results has been provided.

Recent studies have highlighted the growing cybersecurity risks associated with AI and emerging technologies. The study by Radanliev et al. \cite{C5}examines cybersecurity threats, exploits, and vulnerabilities in AI-driven software bills of materials (SBOM) and emphasizes the need for security-aware AI development. While their work focuses on risk identification and policy considerations, our approach enhances IDS transparency through explainability techniques, enabling cybersecurity professionals to better interpret AI-based IDS decisions.

Similarly, another study by Radanliev et al. explores AI security risks in IoT systems, highlighting the impact of adversarial attacks and system vulnerabilities \cite{C6}. Their research identifies key risk factors in AI-driven IoT security but does not address how explainable AI can mitigate these risks. Our work uniquely contributes by integrating LIME and ELI5 to provide both local and global explanations, improving interpretability for security experts using IDSs.

Unlike these risk-oriented studies, our framework enhances the explainability of IDS models, bridging the gap between AI decision-making and human interpretability. By providing clear insights into IDS predictions, we empower cybersecurity professionals to make more informed decisions and mitigate AI security risks proactively.

The use of spatial-temporal features in IDSs has been explored in various domains, including in-vehicle networks \cite{E3}. This study demonstrated the effectiveness of using spatial-temporal representations to enhance intrusion detection. However, these models remain largely opaque, making the integration of explainability techniques, as in L-XAIDS, essential for practical cybersecurity applications.

\subsection{Local Interpretable Model-Agnostic Explanations (LIME)}\label{subsec2}
\quad LIME is an algorithm proposed in \cite{R63} that proposes an implementation of local surrogate models for explainability purposes.  The idea behind LIME's design is to provide explainations to individual predictions by training local surrogate model. 

The usage of explainable methods in the field of intrusion detection or prevention is a very infrequent instance and same is true for LIME as well. 

LIME and ELI5 were chosen over alternative explainability techniques like SHAP and Grad-CAM due to their applicability to tabular data and their ability to provide both local and global explanations in an interpretable manner. SHAP, while powerful, is computationally intensive and requires significant runtime for high-dimensional datasets like UNSW-NB15. Grad-CAM, on the other hand, is primarily designed for convolutional neural networks (CNNs) and is less suited for tabular IDS data. LIME allows for localized feature importance visualization, which is crucial for auditing IDS decisions, while ELI5’s permutation importance provides a global understanding of feature contributions. This combination ensures a balance between computational efficiency and interpretability for IDS applications.

Most of the research citing LIME is focused in the fields of computer vision\cite{R64} \cite{R65} and natural language processing \cite{R66}\cite{R67}. SHAP (Shapely-value) has been used in \cite{R68}\cite{R69} to bring explainbility to the decision making process in the field of biology. Most of the researches based on explainable algorithms do not cater the cybersecurity applications of these algorithms.
In \cite{R70} Amareasinghe and Manic,  implemented  an extension of a technqiue called Layer-wise Relevance Propagation (LRP) \cite{R71} to generate contributions of the input features to help the users to understand what features effect the decision making process in intrusion detection. Similarly, Marino et al. \cite{R72} proposed an adversarial technique to explain the incorrect classifications made during the intrusion detection process. Li et al. \cite{R73} employed a technique called "Local Explanation Method using Nonlinear Approximation''  \cite{R74} to interpret the prediction made by an anomaly-based Intusion Dectection System. 
In terms of explainability, models can be classified into two categories\cite{R75}: 
\begin {itemize}
\item models providing global explanations and,
\item models providing local explanations . 
\end {itemize}
Global explanations refer to the explanations provided to the users to help understanding the overall nature of the model. Whereas, Local explanations refer to just inspecting an input, and trying to figure out why the algorithm made this certain prediction for a specific instance. In this paper, LIME is used to provide local explanations for individual instances and combined with ELI5 we also tried to provide global explanation as well to make the model more explainable.
\subsection{Explain Like I'm 5 (ELI5)}\label{subsec3}
\quad One of the more efficient ways to get an insight of tree based model is by permuting the values of each feature and to inspect the changes one-by-one and evaluating with respect to model performance. This is commonly know as the Permutation Importance method. 
ELI5 is a package that is used to inspect and explain the predictions made by machine learning classifiers. ELI5 is mostly used by debug algorithms such as XGBoost, sklearn regressors, Keras and CatBoost etc. ELI5 can also be used to implement the inspection and explainable aspect of black-box models. TextExplainer is one of the sub-function of ELI5 that enables it to explain predictions of any text classifier using LIME algorithm. Some experimental utilities can also be used for LIME with non-text data and arbitrary black-box classifiers. Wheras, Permutation importance sub-functionality is used to calculate feature significance for black box estimators.

\section{Methodology}\label{sec3}
\quad In this section we propose L-XAIDS framework to improve the explainability of any IDS. When security engineers use IDSs predictions for threat intelligence or SIEM data feeds, explainability of the algorithm is
almost as important as the accuracy. Fig. \ref{fig4} illustrates the design of the proposed structure, which includes two sections. The left section is the depiction of conventional IDS, and the right section is used to illustrate the working of explainability module.
In relation to the standard tasks of IDS, the structure integrates the dataset, established intrusion detection models, and their respective forecasts.

The main aim of this introduced structure is to enhance the understandability of the IDS. As such, local and global explanations are generated to heighten the confidence of cybersecurity experts in the IDS.
The global explanation depicts the significant attributes of the model and illustrate relationships between the value of a feature and the impact on the prediction (detailed discussion in Sec. V (Fig. \ref{fig12} and \ref{fig13})). The local explanations elaborate the predictions being made and the significant attributes effect on the final decision  (detailed  discussion in Sec. V (Fig. \ref{fig14}-\ref{fig16})). 

\subsection{Dataset}\label{subsec1}
The quality of packets captured greatly influence the final predictions of network data feature classification, and it is essential to have a sufficient amount of events to ensure accurate results. The presence of multiple NIDS datasets with varying feature sets presents a challenge, as it is not feasible to extract all the characteristics once the model is deployed. To enhance reliability and increase deployment possibilities, it is crucial to integrate a single feature set across different datasets. The UNSWNB-15 dataset was used in this research and consists of both normal and malicious network traffic samples. It is significant because it contains recent cyber-attacks and realistic network attack data, making it valuable for designing an explainable AI for intrusion detection. UNSW-NB15 includes 49 features with class labels distributed across four CSV files, and contains NaN values and non-uniform class distribution. These factors directly impact the performance and accuracy of any machine learning model.

The UNSW-NB15 dataset was selected due to its diverse attack types, realistic traffic simulation, and balanced normal vs. malicious samples. It includes modern threats such as DoS, botnets, and reconnaissance attacks, making it highly suitable for evaluating IDSs. To assess the generalizability of L-XAIDS, preliminary tests were conducted on the CICIDS2017 dataset, which yielded similar feature importance rankings, reinforcing the model’s adaptability across datasets. Future work will extend this evaluation to additional cybersecurity datasets.

The dataset underwent several preprocessing steps to ensure data integrity. Missing values were handled by removing records with excessive NaN values (>30\%) and imputing missing numerical values using median imputation to prevent data leakage. Categorical features were label-encoded for compatibility with machine learning models. To address class imbalance, SMOTE (Synthetic Minority Over-sampling Technique) was applied with a 1:1 ratio for normal and attack traffic. We used a k-nearest neighbors (k=5) approach to generate synthetic samples while avoiding overfitting. These preprocessing steps enhance the generalizability and robustness of our model in real-world settings.

\subsection{Feature Selection}\label{subsec2}
Feature selection refers to identifying which features have a substantial impact on the outcome. This can lead to reduced model overfitting, processing time, and improved accuracy. One such technique is using Global explanation algorithms such as ELI5 (Explain Like I'm5) and scikit-learn to identify the attributes that have significant impact on the outcome of the prediction. We have paired up the features extracted from global explanations with correlation heatmap, which displays the relationship between the features. If a group of features has a high correlation, one feature from the group can replace or eliminate the others.  According to the heatmap in Fig.\ref{fig7}, it is evident that a substantial relationship exists among the features such as (sloss, sbytes), (dloss, dbytes), (sttl, ct state ttl) and others. 
The second stage of feature selection involved considering only the most significant features extracted by the use of ELI5 and scikit-learn for global explanations as shown in Fig. \ref{fig12} and \ref{fig13}.

Algorithm 1 illustrates the process to identify significant features.  The "Identifying significant features" algorithm aims to discern the most influential attributes or features that contribute to a model's prediction decisions. The approach harnesses the power of the LIME (Local Interpretable Model-agnostic Explanations) framework to achieve this.

Initially, the algorithm takes as its input a trained model, a training dataset (both features X\_trn and labels Y\_trn), and a comprehensive list of all available features in the dataset (All\_Features). The first task the algorithm embarks upon is to set up an explainer, Expl, using the LIME framework, making it suited for the given training data and feature set. With this explainer, the algorithm aims to elucidate the model's predictions by focusing on localized interpretations.

The algorithm has a predefined threshold set for the number of significant features it aims to process, which is 10 in this context (N← 10). It then begins a loop to iterate through each Normal instance. While it's not explicitly defined what Normal refers to, it can be inferred that the algorithm is looking at typical or standard instances. For each of these standard instances, referred to as nml, the algorithm fetches the respective data from the training set and uses the LIME explainer (Expl) to provide an explanation for the model's predictions on that instance. The aim is to understand which features have the most bearing on the prediction for that particular instance.

Once the loop is complete and explanations have been generated for all the Normal instances, the algorithm proceeds to extract the significant features from these explanations. This extraction function, ExtractFeatures(), presumably aggregates and assesses the importance scores across all the instances to pinpoint the top features influencing the model's decisions.

In the end, the algorithm returns SigF, a collection of these significant attributes. By identifying these crucial attributes, users or developers can prioritize their focus on these features, be it for feature engineering, interpretation, or subsequent analyses, ensuring better model clarity and efficiency.

\begin{algorithm}
\caption{Calculate $y = x^n$}\label{algo1}
\begin{algorithmic}[1]
\Require $n \geq 0 \vee x \neq 0$
\Ensure $y = x^n$ 
\State $y \Leftarrow 1$
\If{$n < 0$}\label{algln2}
        \State $X \Leftarrow 1 / x$
        \State $N \Leftarrow -n$
\Else
        \State $X \Leftarrow x$
        \State $N \Leftarrow n$
\EndIf
\While{$N \neq 0$}
        \If{$N$ is even}
            \State $X \Leftarrow X \times X$
            \State $N \Leftarrow N / 2$
        \Else[$N$ is odd]
            \State $y \Leftarrow y \times X$
            \State $N \Leftarrow N - 1$
        \EndIf
\EndWhile
\end{algorithmic}
\end{algorithm}

\begin{algorithm}[H]
\caption{Identifying Significant Features}
\label{alg:identifying_features}
\begin{algorithmic}[1] % The [1] makes the line numbers start from 1

  \Require Model, $X_{trn}$, $Y_{trn}$, All\_Features
  \Ensure Significant attributes: SigF

  \State $Expl \leftarrow {Lime}(X_{trn}, Y_{trn}, {All\_Features})$
  \State $pred \leftarrow$ Model’s prediction function
  \State Significant features to process: $N \leftarrow 10$
  \For {each normal sample stored in $X_{trn}$}
    \State explanation[nml] $\leftarrow {Expl}(X_{trn}[nml], pred, N)$
  \EndFor
  \State $SigF \leftarrow {ExtractFeatures}(explanation)$
  \Return $SigF$
\end{algorithmic}
\end{algorithm}

\subsection{XAI pipeline}\label{subsec3}
The machine learning process involves loading data, preprocessing it, and implementing an AI model. We have used three traditional supervised learning algorithms (DT, MLP, XGBOOST) and the following describes each step in the implemented method.

Data Loading: Bringing the raw data into the system is the first step for any ML analysis technique. Raw data can comprise of dataset of log files or a database. UNSW-NB15 dataset includes four files and they were combined into one single file. The Pandas library was used to load the data, using functions from sci-kit-learn.

Data Preprocessing: To standardize the UNSW-NB15 dataset we used StandardScaler that recognizes the "garbage in, garbage out." Null values were removed from the data frame using the drop.null() function. The dataset was large and contained many null values, so this was necessary. We have used feature encoding, scaling as well as label encoding to handle null values present in the dataset and converting the categorical data in numerical form with the help of label encoder.  

Train test split: The data was split into 80-20 for training and testing purposes respectively using sklearn model selection library was used. The data was also balanced using an imblearn Randomundersampler and the SMOTE synthetic minority oversampling technique.

K-fold cross-validation with k = 5 was used for training data, which involves splitting the training data into five parts and training the algorithm multiple times. This approach allowed for tuning the parameters and preventing overfitting.

XAI is an approach to building machine learning models that are transparent, interpretable, and explainable. In the context of Intrusion Detection Systems (IDS), an XAI pipeline can help provide insights into how the model makes decisions, which is critical for identifying and responding to potential threats.

One approach to building an XAI pipeline for IDS is to use a combination of Lime, Eli5, decision trees, MLP, and XGBoost with k-fold cross-validation of 5. Lime and Eli5 can be used for local interpretability, which provides explanations for individual predictions. Decision trees and MLP are both interpretable models that can be used for global interpretability, which provides an overall understanding of the model's behavior. XGBoost can be used as a high-performance machine learning model, which can be further interpreted using Lime or Eli5.

K-fold cross-validation of 5 can be used to evaluate the performance of the model and ensure that the pipeline is robust and generalizable. Overall, this XAI pipeline can help provide a better understanding of how an IDS model is making decisions, which can help improve its accuracy and reliability in detecting potential threats.

Algorithm 2 illustrates the working of L-XAIDS as a whole from loading and preprocessing to generating explanations for local and global instances.. The L-XAIDS algorithm is aimed at achieving an explainable intrusion detection system (IDS) using the UNSW-NB15 dataset. The overall goal is not just to make predictions about possible intrusions but also to understand the reasons behind those predictions to enhance both the system's efficacy and transparency. Here's a breakdown of its operation:
\begin{itemize}
\item The algorithm starts by loading and preprocessing the UNSW-NB15 dataset, a comprehensive dataset used for building and evaluating intrusion detection systems.
\item It then splits this dataset into two parts: a training set for training the classifiers and a testing set for evaluating their performance.
\item A set of classifiers, such as Decision Tree, are initialized. These classifiers will be individually trained and tested.
\item For each classifier, the training process begins, followed by the prediction phase on the testing set. 
\item To understand the underlying decision-making of the classifier, the algorithm employs explainers. Two primary explainers are highlighted here: LIME and ELI5.
\item LIME (Local Interpretable Model-agnostic Explanations) provides local explanations, shedding light on why a specific instance was classified in a particular manner. For every classifier, the algorithm creates a LIME explainer, selects an instance from the testing set, and generates explanations for it. These explanations are then visualized and interpreted to understand the classifier's decision-making process for that specific instance.
\item ELI5 (Explain Like I'm 5) is another explainer that offers global and local feature importances. Global importances give an understanding of which features generally influence the classifier's decisions the most, while local importances focus on a specific instance's classification. After displaying these importances, the results are interpreted to comprehend the classifier's reasoning.
\item Steps 4 through 7 are iteratively repeated until the desired level of explainability is achieved.
\item Finally, the algorithm returns the explainable IDS features, ensuring that the IDS is not only accurate but also interpretable.
\end{itemize}

\begin{algorithm}[H]
\caption{L-XAIDS Algorithm}
\begin{algorithmic} 
    \Require UNSW-NB15 dataset
    \Ensure Explainable IDS features

    \State \textbf{1.} Load and preprocess the UNSW-NB15 dataset
    \State \textbf{2.} Divide the dataset into training and testing sets
    \State \textbf{3.} Initialize a list of classifiers, e.g., [RandomForest, LogisticRegression, DecisionTree]
    
    \For {each classifier in the list of classifiers}
        \State \textbf{4.} Train the classifier on the training set
        \State \textbf{5.} Predict labels for the testing set
        \State \textbf{6.} Calculate classification performance metrics
        \State \textbf{7.} Initialize a list of explainers
        
        \For {each explainer in the list of explainers}
            \If {explainer = LIME}
                \State \textbf{8.} Create a LIME explainer object for the trained classifier
                \State \textbf{9.} Choose an instance from the test set to explain
                \State \textbf{10.} Generate explanations for the chosen instance
                \State \textbf{11.} Visualize and interpret the explanations
            \ElsIf {explainer = ELI5}
                \State \textbf{12.} Display global feature importances for the trained classifier using ELI5
                \State \textbf{13.} Display local feature importance for an instance using ELI5 (if applicable)
                \State \textbf{14.} Interpret the results
            \EndIf
        \EndFor
        
        \State \textbf{15.} Based on insights from the explanations, modify feature set or the classification algorithm
        \State \textbf{16.} Retrain the classifier and evaluate its performance
        \State \textbf{17.} Repeat Steps 4-16 until explainability is achieved
    \EndFor

    \Return Explainable IDS features
\end{algorithmic}
\end{algorithm}

\subsection{Classification and Regression Tree}\label{subsec4}
The decision tree is a widely used and highly effective tool for classification and prediction. The result of testing one or more attributes can be represented by a decision tree, depending on its purpose. In the tree structure, the inner nodes symbolize tests, the branching nodes indicate the results, and the terminal or leaf nodes denote classes.

This technique enables individuals or organizations to weigh the costs, possibilities, and benefits of each action. They can be used to compare potential outcomes, either by developing an algorithm to calculate the best option mathematically or by being used in informal discussions. The tree typically starts with one node that branches out into possible outcomes, forming its arborescent structure. There are three types of nodes in a decision tree: likelihood node, call node, and finish node.

When using decision trees, some assumptions are made, such as considering the entire training set as the root, preferring categorical values for feature values, and distributing records based on attribute values through recursive splitting. Moreover, statistical techniques are employed to classify root attributes based on their significance.
Building a decision tree involves iteratively asking questions to divide the sample points. The maximum depth parameter specifies and regulates the tree's depth, halting the model's split once this limit is achieved. The optimal value for maximum depth is dependent on the specific data set. Typically, a shallow tree height might heighten the risk of overfitting, and crucial data pertinent to the dataset could potentially be overlooked.

\subsection{Local Explanation (LIME)}\label{subsec5}
\quad The LIME algorithm provides explanations for the predictions made by a machine learning model by determining the values of significant attributes and their contribution to the final prediction. This method helps security experts understand the reasoning behind the decisions made by an intrusion detection system (IDS) based on specific data.
To achieve this, the LIME algorithm alters the occurrence of the data several times by making small changes in the values, and creates a linear model using the fake data around the altered observation. The prediction of the altered data is then compared with the original data to measure the distance between them, and a similarity score is calculated based on the distance. The best 'm' features are selected to represent the predictions from the altered data and a simple model is fit using these selected features.
The following is the outline for building the pipeline model in the LIME algorithm:
\begin {itemize}
\item IMPORT necessary libraries and the dataset
\item PRE-PROCESS the data to handle missing, NaN and infinity values.
\item SCALE the data
\item ASSIGN names to the machine learning models
\item FOR every individual model:
\item Store the model selection outcomes using 10 splits
\item Determine the results via cross-validation scoring methods
\item Append the new and existing results
\item Display the model accuracy and classification report
\item END LOOP
\end {itemize}

The mathematical equation for Local Interpretable Model-Agnostic Explanation (LIME) is as follows:
Let $X$ be the feature vector of a given instance, and let $f$ be a machine learning model that predicts the output for $X$. The goal of LIME is to provide a local explanation for the prediction of $f$ at $X$, in terms of the feature importance scores of the input features. LIME approximates the complex model $f$ with a simpler, interpretable model $g$, such as a linear model, within a small neighborhood of $X$. The neighborhood is defined by a kernel function $K$ that assigns weights to the training instances based on their proximity to $X$. 
The local approximation $g$ is trained on a set of perturbed instances $S$, where each instance in $S$ is generated by randomly perturbing the original instance $X$ while keeping its label fixed. The perturbations can be made in various ways, such as adding noise or deleting features. The feature importance scores of the input features are computed based on the weights assigned by $g$ to each feature. The higher the weight of a feature, the more important it is in predicting the output for $X$.

Formally, the explanation provided by LIME can be expressed as follows:

\begin{equation} \label{eq1}
LIME(X) = argmax\_g L(f,g,K)+\sigma(g) , 
\end{equation}
where $L$ is a loss function that measures the dissimilarity between $f$ and $g$ on the training instances in $S$, and $\sigma$ is a complexity penalty that discourages complex models. The argmax operator selects the model $g$ that maximizes the tradeoff between accuracy and interpretability. The feature importance scores can be computed based on the coefficients of $g$ or the weights assigned to the input features by $g$. LIME provides local explanations for individual instances, allowing partial understanding of the complex model, but not a global understanding.

\begin{algorithm}[H]
\caption{Algorithm for Generating LIME Explanation}
\begin{algorithmic} 
    \Require Pretrained classifier (f\_Model), Instance (x), Similarity Kernel K\_x
    \Ensure LIME Explanation (L)
    
    \State \textbf{1.} Initiate $C \leftarrow \{\}$
    \For {each index $ind$ in $\{1, 2, 3, \ldots, N\}$}
        \State \textbf{2.} {Generate a sample c\_0 around the instance (x)}
        \State $c_0 \leftarrow (x_0)$
        \State \textbf{3.} {Accumulate the sample data in cluster C}
        \State $C \leftarrow \{(c_0, f\_{{Model}}(c_0\_index), K_x(c_0\_index))\}$
        \State \textbf{4.} {Perform Linear Regression with c0\_index as features, $f_{{Model}}(c_0\_index)$ as target}
        \State $L \leftarrow {LinearRegression}(C, N)$
    \EndFor
    \State \textbf{5.} Generate Explainer
    \State LIME Tabular Explainer($X_{trn}$, features, class\_names=\'NORMAL\', \' MALICIOUS\', feature\_names\_cat)
    \Return $L$
\end{algorithmic}
\end{algorithm}

The algorithm 3 takes the following inputs for generating local explanations:

\begin{itemize}
\item A trained classifier model f
\item An instance x and its perturbed datapoints x\_0
\item A dataset Dset with N samples
\item A similarity kernel SKx
\item The output of the algorithm is a local explanation of the black-box model's prediction for the given instance x using LIME (L).
\end{itemize}
The method of the algorithm involves the following steps:

Initialize an empty cluster C.
For each index ind in the set \{1, 2, 3, ..., N\}:
a. Sample perturbed data around the given instance x\_0 to obtain a perturbed instance c\_0,ind.
b. Add the perturbed instance c\_0,ind, the predicted output of the black-box model f for c\_0,ind, and the similarity kernel SKx(c\_0,ind) to the cluster C.
Use the perturbed instances in cluster C as features and the corresponding predicted outputs of the black-box model f as the target to train a linear regression model L that explains the black-box model's behavior in the local region around the given instance x.
Return the trained linear regression model L as the local explanation for the black-box model's prediction for the given instance x.
In the LIME (Local Interpretable Model-agnostic Explanations) method, perturb data points and similarity kernels are used to generate local explanations of a black-box model's predictions.

Perturb data points refer to modified instances of the original data point that are created by perturbing some of its features while keeping the others fixed. These perturbed instances are used to estimate the local behavior of the black-box model around the original data point.

The perturbations can be done in various ways, such as adding Gaussian noise, replacing a feature value with a random value from the feature's distribution, or discretizing continuous features. The goal is to create perturbed instances that are similar to the original instance, but different enough to cover a diverse range of possible scenarios.

The similarity kernel is a function that measures the similarity between two instances, such as the Euclidean distance or cosine similarity. It is used to weigh the contribution of each perturbed instance in the local explanation, based on how similar it is to the original instance.

For each perturbed instance, the similarity kernel is applied to measure its distance from the original instance. The weights of the perturbed instances in the local explanation are then computed using the similarity kernel, with closer instances having higher weights. This helps to focus the local explanation on the most relevant perturbed instances that are similar to the original instance.
The left section showcases the likelihood of predictions. Thirteen features are highlighted in the central section. The process of binary classification is visualized in two distinct colors, blue and orange, representing class 0 and class 1 attributes respectively. Horizontal bars demonstrate the comparative significance of these features, represented by floating-point numbers. This color scheme is maintained across all sections. Also included is a compilation of actual values for the top 13 factors.

\subsection{Global Explanation (ELI5)}\label{subsec6}
\quad ELI5 (Explain Like I'm 5) is used to generate global explanation for the IDS predictions. It facilitates the evaluation of machine learning classifiers by detailing their predictive processes. This is achieved through the allocation of weights to specific decisions, the generation of decision trees, and the presentation of the importance of various features of the model provided to the tool. It also explains the importance of the features in a table format as shown in Figure \ref{fig12}.

In Figure \ref{fig12}, we employ the show\_weights() function from ELI5 to display feature significance in descending order for UNSW-NB15. This can be utilized to understand how a model functions on a broad scale. As depicted in the figure, the feature 'sttl' carries the most substantial approximate weight of 0.2755, whereas the feature 'ct\_dst\_ltm' holds the least importance, with an estimated weight of 0.0005 before the feature weights diminish to zero. We can also note that the 'sttl', 'ct\_dst\_sport\_ltm', and 'sbytes' features are marked in green, while features from 'sjit' and 'dinkpkt' onwards are not. The green marker indicates features that have a significant and positive contribution to the model.
The value after the ± sign measures how performance varied from one-reshuffling to the next.

Algorithm 4 illustrates the generation of ELI5 permutation importance for global instances, importance values and feature weights that is used as input for generating global explanations in algorithm 3.

The pseudo-code in algorithm 4 is generating a global explanation for a given instance (x) in the dataset using permutation importance and feature weights.

First, the code predicts the class label for the instance using the pretrained classifier (f\_Model). Then, it calculates permutation importance using the PermutationImportance function from ELI5, which randomly shuffles the values of each feature in the instance, calculates the change in the prediction accuracy, and records the feature importance scores. The permutation importance is calculated N times for stability and reproducibility.

Next, the code obtains the feature weights using the show\_weights() method from ELI5, which calculates feature importance scores using various methods such as permutation importance, SHAP values, etc. Then, the code calculates the average permutation importance score for each feature by taking the mean of the permutation importance scores obtained in the previous step.

Finally, the code combines the feature weights and the average permutation importance scores for each feature to generate the global explanation in the form of a dictionary where each feature is associated with its weight and permutation importance. The global explanation is then returned.

\begin{algorithm}[H]
\caption{Algorithm for Generating ELI5 Permutation Importance}
\begin{algorithmic}
 \Require Pretrained classifier \( f_{{Model}} \), Instance \( x \), Number of iterations \( N \)
 \Ensure ELI5 permutation importance values, feature weights, and performance variance from shuffling

 \State 1. \( y_{{pred}} = f_{{Model}}.{predict}(x) \) \Comment{Calculate permutation importance}
 \State 2. \( {perm} \gets {PermutationImportance}(f_{{Model}}, {scoring} = 'accuracy', n_{{iter}} = N, {random\_state} = 0) \)
 \State 3. \( {perm.fit}(x, y_{{pred}}) \)
 \State 4. \( {weights} \gets {eli5.show\_weights}(f_{{Model}})\) \Comment{Get feature weights} 
 \State 5. \( {avg\_perm\_imp} \gets \{\}   \)
 \State 6. \( {feature\_names} \gets {perm.features\_} \)
 \For{i = 0 \textbf{to} {len(feature\_names)} - 1}
    \State \( {feature\_name} \gets {feature\_names}[i] \)
    \State \( {feature\_imps} \gets {perm.feature\_importances\_}[i] \)
    \State \( {avg\_perm\_imp[feature\_name]} \gets {feature\_imps.mean()} \) \Comment{Compute average permutation importance}
 \EndFor
 \State 7. \( {global\_explanation} \gets \{\} \) \Comment{Combine feature weights and average permutation importance}
 \For{{feature} \textbf{in} {feature\_names}}
    \State \( {global\_explanation[feature]} \gets \{ 'weight': {weights[feature]['weight']}, 'perm\_imp': {avg\_perm\_imp[feature]} \} \)
 \EndFor
 \State 8. \textbf{Return} \( {global\_explanation} \)
\end{algorithmic}
\end{algorithm}

\subsection{Linking Feature values with attack traffic}\label{subsec7}
\begin{figure}[!t]%
\graphicspath{ {./pictures/} }
\centering

\includegraphics[width=1\textwidth]{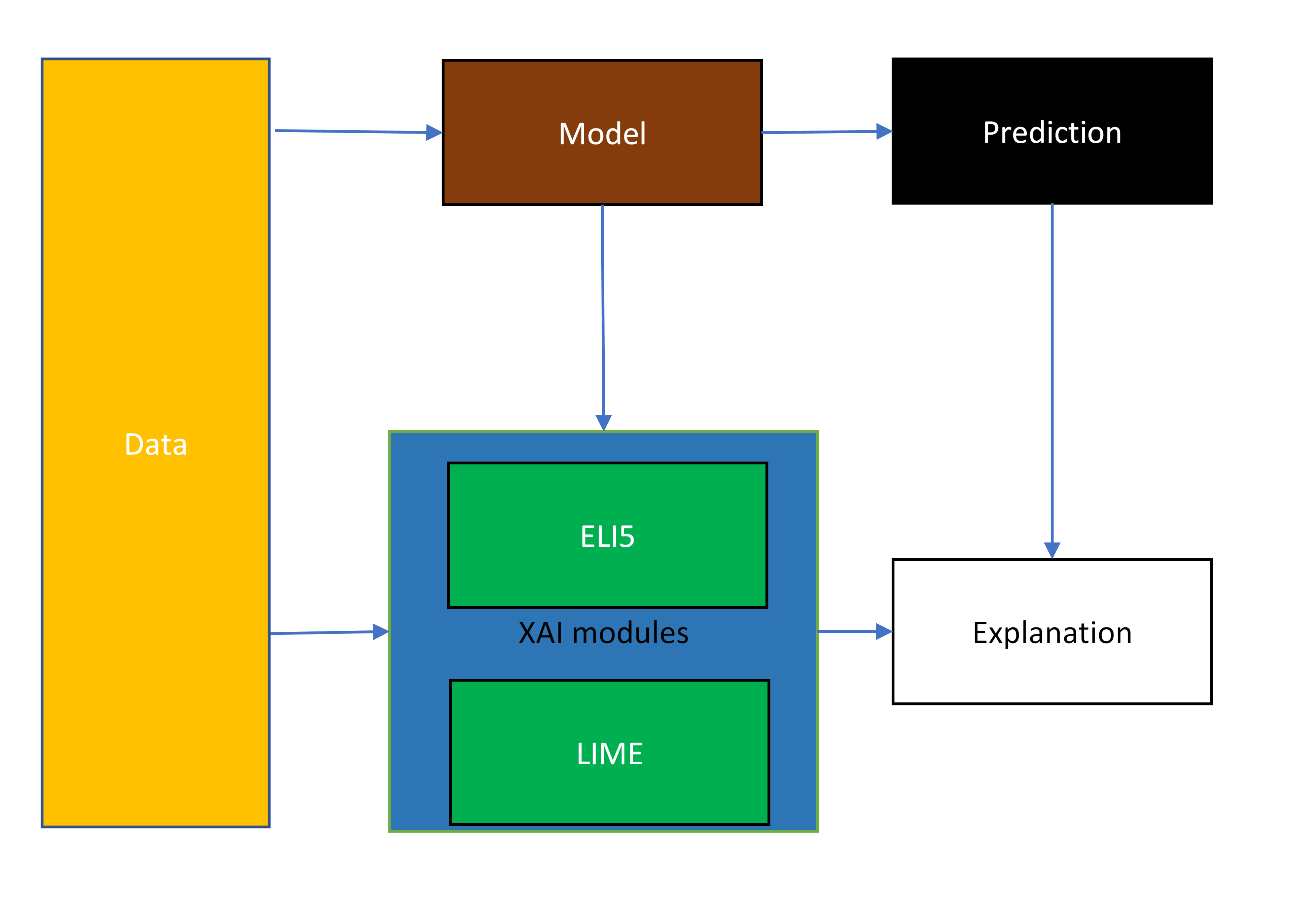}
\caption{Overview of how to use LIME/ELI5 for explainable model}\label{fig1}

\end{figure}

\begin{figure}[!t]%
\graphicspath{ {./pictures/} }
\centering

\includegraphics[width=0.6\textwidth]{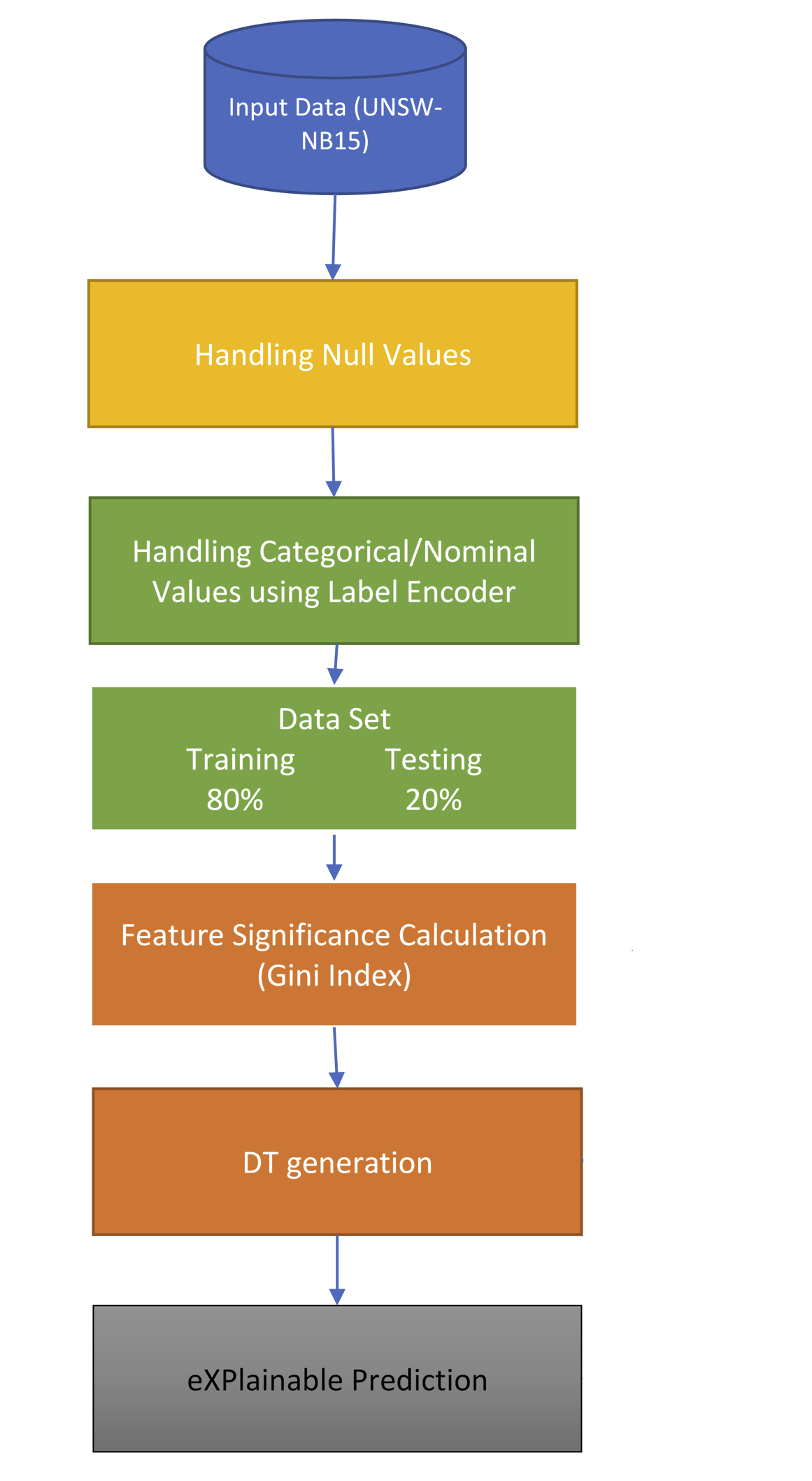}

\caption{Flow of proposed algorithm}\label{fig2}

\end{figure}

Having a basic understanding of the key features of the model is not sufficient to grasp the essence of malicious traffic. To have a deeper understanding, it is crucial to examine the connections between the feature values and the predictions made by intrusion detection systems (IDSs). This can aid cybersecurity experts in gaining a better comprehension of the model. Understanding the relationships between feature values and different forms of attacks will allow cybersecurity engineers to gain a more extensive understanding of both the IDS and network breaches.
Fig.\ref{fig1} illustrates the high-level architecture of the model proposed for explainable IDS.  This technique employs the strategy of connecting the two modules, LIME and ELI5. Each ELI5 value generates explanation of individual contribution of each feature in the model and is significant for the overall decision making process. 
LIME, on the other hand, weighs the instances in the regression model with respect to their closeness to the original instance. 
Fig.\ref{fig2} outlines the six-step procedure for Lime-Explainable AI for IDS.
First step is preparation of dataset that comprises of feature encoding and scaling followed by handling of null values and converting the categorical data in numeric form. 
Second step of the algorithm is to handle the categorical and nominal values using Label encoder.
Third step is splitting the dataset in 80-20 split for training and testing sets respectively.
Fourth step is to generate feature significance matrix via Gini Index and populating ELI5 for global level explanation of the dataset.
Fifth step is generating a Classification and Regression Tree (CART) followed by a LIME based local level explanation for packets under analysis.
The working of this algorithm and all the subsequent steps has been discussed in detail in Experimental details.

Algorithm 5 identifies intrusions via predictions and combining them with local and global explanations, along with defining thresholds and what to do if certain threshold doesn't meet.
Algorithm 5 is designed to identify potential intrusions in network data by not just relying on predictive decisions but also leveraging explanations to provide insights into these decisions. Its core functionality revolves around the intricate balance between automated predictions and explainability to enhance the intrusion detection process's accuracy and interpretability.

Initially, the algorithm accepts as input a set of network data, ND1, and a set of attributes, SetF. The first step involves using the L-XAIDS algorithm (from the previous discussion) on the network data to get a prediction value. This prediction can be seen as a score, perhaps representing the likelihood of the data being malicious.

The crux of the algorithm lies in its conditional assessment of this prediction value. If the prediction value is less than or equal to 80 (the exact metric of this value isn't given but can be inferred as a threshold of classification certainty or risk assessment), the data is subjected to further scrutiny. Instead of immediately classifying it as "Malicious" or "Normal," the algorithm seeks more clarity by generating explanations for its preliminary prediction using LIME and ELI5 – two popular tools for model interpretability. Thus, for the given network data, both lime\_explanation and eli5\_explanation functions are invoked to provide detailed rationales for the model's decision.

After generating these explanations, the algorithm consolidates them and associates them with the network data, ND1, in the explanations dictionary. The output, in this case, will be a warning generation indicating potential malicious activity, accompanied by the LIME and ELI5 explanations. These explanations will allow network administrators or users to understand why a particular piece of data was deemed potentially malicious, offering more granular insights into the decision-making process.

On the other hand, if the prediction value exceeds 80, the algorithm perceives the network data as non-threatening, promptly returning "No Warnings." This bifurcated approach ensures that only ambiguous or potentially risky data undergoes a deeper interpretability analysis, thereby optimizing computational resources and enabling timely intervention when required.

\begin{algorithm}[H]
\caption{Identifying Intrusions using Decisions and Explanations}
\begin{algorithmic}
 \Require Network data \( ND1 \), attributes \( SetF \)
 \Ensure Malicious/Normal

 \State \( Prediction \gets {L-XAIDS}(ND1) \)
 \If {Prediction value \( \leq 80 \) (Malicious or Normal)}
    \State \textbf{Generate LIME and ELI5 Explanations}
    \State \( {exp.lime} \gets {lime\_explanation}(ND1) \)
    \State \( {exp.eli5} \gets {eli5\_explanation}(ND1) \)
    \State \( {explanations}[ND1] \gets {Expln}(ND1, {pred}, N, {exp.lime}, {exp.eli5}) \)
    \Return Warning Generation, \( {exp.lime} \), \( {exp.eli5} \)
 \Else
    \Return No Warnings
 \EndIf
\end{algorithmic}
\end{algorithm}

\begin{figure}[ht!]%
\graphicspath{ {./pictures/} }
\centering

\includegraphics[width=0.8\textwidth]{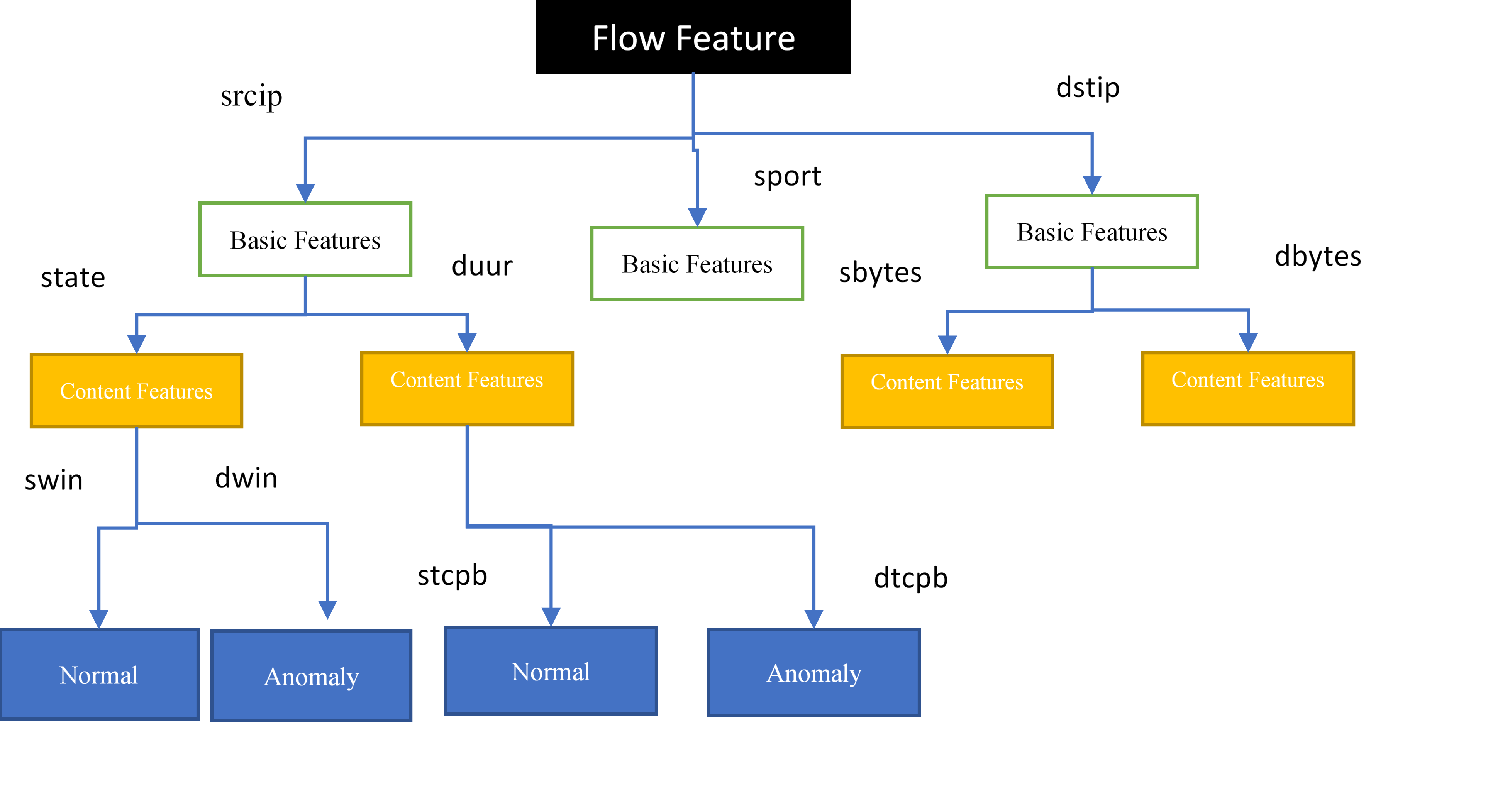}
\caption{Flow of L-XAIDS CART Tree}\label{fig3}

\end{figure}

\begin{figure*}[!t]%
\graphicspath{ {./pictures/} }
\centering
\includegraphics[width=0.9\textwidth]{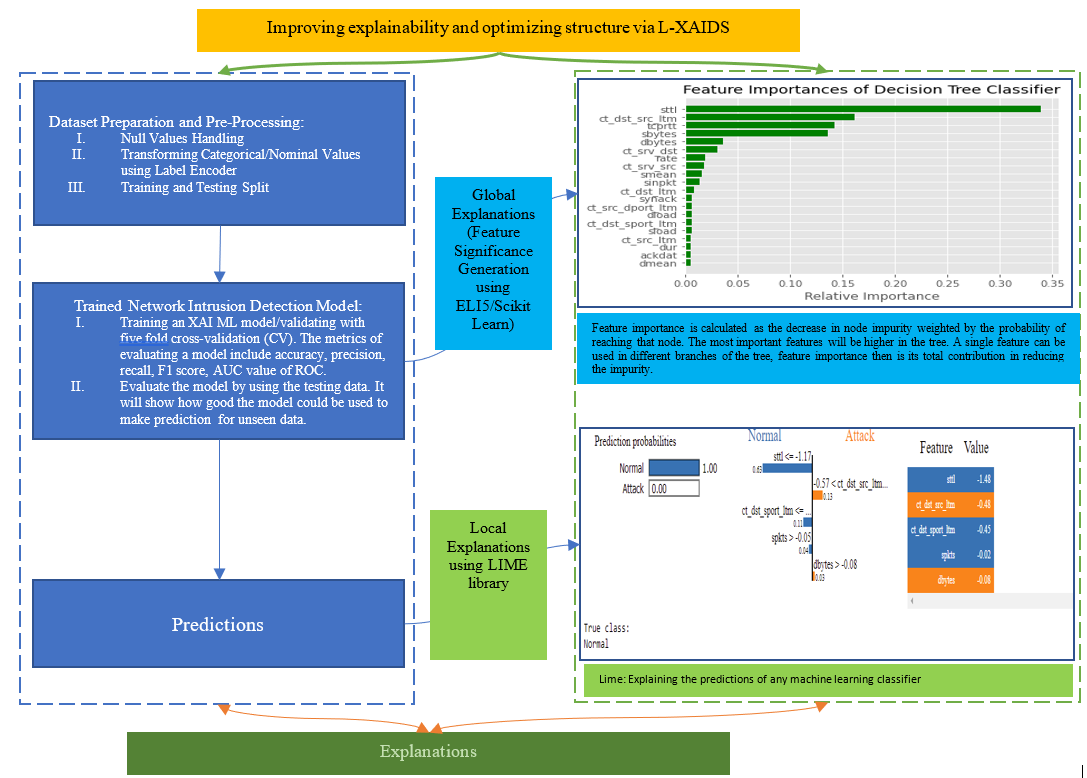}
\caption{Structure of proposed : L-XAIDS framework}\label{fig4}
\end{figure*}

The framework proposed in this section will assist in improving the transparency of the IDS system by providing explainable rationals of the decisions made. This will enable humans-in-the-loop to understand the working of the model. Fig. \ref{fig1}, \ref{fig2} and \ref{fig3} illustrate the detailed working of the framework proposed. 
Fig. \ref{fig3} shows the flow of proposed algorithm based on the features categories.  Branching out features starting from Flow feature, basic features, content features and consequently labeling a packet as normal or anomalous/malicious. 
The hierarical structure of the model comprises of two distinct modules as shown in Fig. \ref{fig4} The left hand side of Fig. \ref{fig4} is the conventional design for the IDS, whereas the right hand side is the explainable modules that improves the transparency of the decision making process of the IDS. The dataset is trained for instrusion detection and global explanation is generated showing the most important and significantly weighted features of the dataset. The local explanations are generated for individual packets showing which features contributed towards labeling of packet as malicious or normal. For example, in Fig. 4 in global explanation sttl, ct-dst-src-Itm, tcprtt, sbytes and dbytes are the five most significant features identified. In the local explanations section, the prediction is explained given these features importance. 
\begin {itemize} 
\item sttl $<=$ - 1.17 is considered normal and the current value -1.48 is significantly lower than the threshold defined, hence 0.63 to the total score of the packet and labeled normal. 
\item The threshold defined for ct-dst-src-Itm is $>$ -0.53 is considered attack and the value for this feature is -0.48, hence contributed 0.13 to the total score of the prediction and labeled as attack traffic.  
\item The threshold defined for ct-ftp-cmd $<=$ - 0.09 is considered malicious and the current value -0.09 meets the threshold defined, hence contibuting 0.12 to the total score of the packet and labeled normal. 
\item The threshold defined for ct-dst-sport-Itm is $<=$ -0.30 is considered normal and the value for this feature is -0.45, hence contributed 0.11 to the total score of the prediction and labeled as normal traffic. 
\item The threshold defined for is-ftp-login is $<=$ -0.09 is considered attack and the value for this feature is -0.09 and meets the threshold, hence contributed 0.05 to the total score of the prediction and labeled as attack traffic.  
\end {itemize}
\quad The conventional IDS module consists of the dataset, the models trained for intrusion detection, and their corresponding predictions. However, the main objective of this research is to bring explainability to the decision making process. Hence, the local and global explanations are provided as part of the algorithm on the right hand side to improve the overseers' trust in the decision making process. ELI5 is used as the method of global explanations in this presented framework. This algorithm can identify the most significant features of the dataset and also exhibits an interpretable relationship matrix corroborating the significance score of a feature and its impact on the local prediction. For local explanations, we have used LIME to explain the prediction of an Intrusion Detection System.

Consequently, the cybersecurity engineers will have an option to validate the predictions made by intrusion detection system by studying and inspecting the local and global interpretations provided. 
\begin {itemize}
\item Local Explanations
LIME can compute contributing weight of all the attributes and provide explanations to the predictions made by the model. These LIME values illustrate each feature's  contribution to the final prediction of the packed under evaluation. For instance, higher sttl rate of a packet may have a higher possibilty of being flagged as an anomalous packed.
\item Global Explanations
ELI5 provides a matrix of significant features and their corresponding contributing values to the model. And this cam be used to picture the overall nature of the model as well as specifying the most relevant features for the decision making on the individual level too. Next step in the process is to sort the features in the decreasing order with respect to the feature significance score, and then the top N most important attributes as specified by the overseer or researcher will be the only ones contributing to the further decision making process.
\item Relationship between feature values and classification of packets
Just specifying the most significant features of the dataset is not enough explaination for the cybersecurity engineers studying the IDS model. Therefore, some sort of metric should be provided that explains the relationship between feature values and predictions made in a quantifying manner. Hence, we first divide the value of feature in several intervals followed by calcuation of LIME values of these features and average them for each interval. 

\end {itemize}

\section{Experiment Details}\label{sec4}

\begin{figure}[!t]%
\graphicspath{ {./pictures/} }
\centering

\includegraphics[width=1\textwidth]{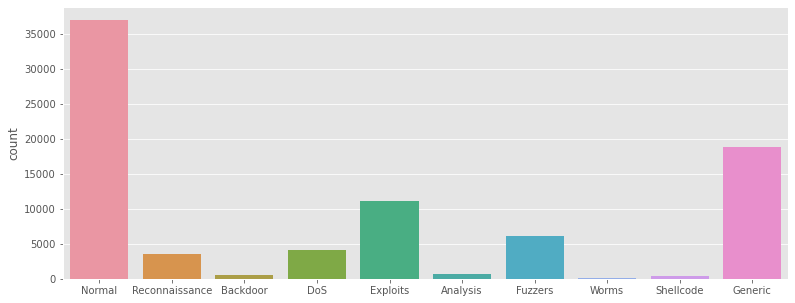}
\caption{Attacks distribution in training dataset}\label{fig5}

\end{figure}

\begin{figure}[!t]%
\graphicspath{ {./pictures/} }
\centering

\includegraphics[width=0.75\textwidth]{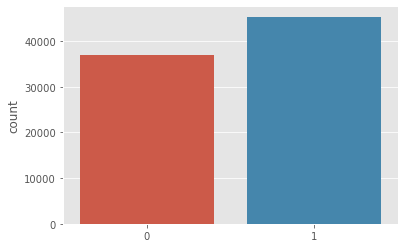}
\caption{Binary level labeling of the traffic from the UNSW-NB15 dataset}\label{fig6}

\end{figure}

\quad UNSW-NB15 is a network traffic data set with different categories for normal activities and malicious attack behaviors from botnets (through classification of attack type including Analysis, Fuzzers, DoS, Backdoors, Generic, Exploits, Shellcode, Reconnaissance and Worms). The dataset consists of 49 features that are numeric in nature.  Fig. \ref{fig5} provide the details and the values distribution of each attack class within the data subsets, whereas in Fig. \ref{fig6} '0' represents Normal and '1' represents Attack behavior. We could see that the dataset is adequately balanced for the binary response variable of activity behavior.

The dataset is trained on three supervised ML binary classifiers, namely Decision Trees, Multi-layer Perceptron Neural Network, and XGBoost. The objective feature will be a binary classification of normal or malicious behavior. Once the models are trained, they will be tested using the processed UNSW-NB15 testing dataset, and their accuracy score will be used to evaluate their performance. The classifiers will not be tuned using hyperparameters. The Decision Trees Classifier and Multi-layer Perceptron Classifier will be implemented using Scikit-Learn, while the XGBoost Classifier will be developed using the XGBoost library.

\begin{figure}[!t]%
\graphicspath{ {./pictures/} }
\centering

\includegraphics[width=0.75\textwidth]{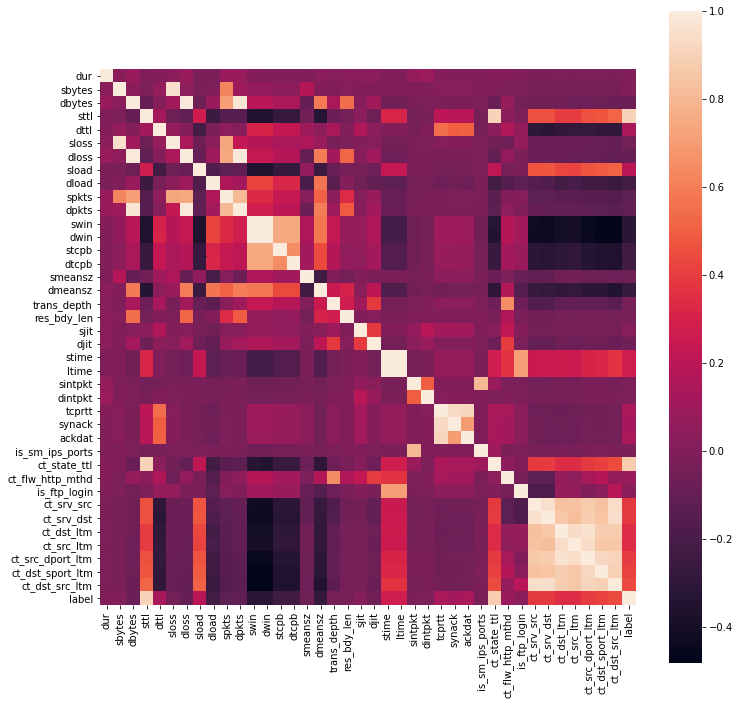}
\caption{Heatmap of UNSW-NB15 features}\label{fig7}

\end{figure}

After training and testing the classifiers, the study will focus on developing interpretable diagrams, feature importance plots, and classification/prediction explanation visuals based on the trained classifiers to detect network traffic behavior in the testing set. To make the ML classifiers explainable, the researchers will explore and modify the following Python packages:

\begin{enumerate}
\item ELI5, which is a visualisation library that is useful for debugging machine learning models and explaining the predictions they have produced.
\item LIME (local interpretable model-agnostic explanations) is a package for explaining the predictions made by machine learning algorithms.

\end{enumerate}
 These packages assist in generating visual explanations of the classifiers' predictions, providing cybersecurity analysts with a deeper understanding of the decision-making procedures within the models..
 \quad Feature selection is of utmost importance for both detection accuracy as well as explainability module. Table 1 illustrates the details of the features included in the dataset and their value types. 
Wherease, Fig. \ref{fig5} illustrates the attack categories included in the dataset. 
Fig. \ref{fig7} depicts the correlation matrix of the features and helps the researchers to understand the correlation relationship between several features in the dataset. For instance, features sttl and dttl have a high correlation value.
Table 2 illustrates the distribution sample of the dataset

\begin{table}[ht]
\centering
\caption{Test Table}
\label{tab:test}
\begin{tabular}{ p{0.5cm}  p{4.5cm}  p{0.5cm}  p{4.5cm}}
\toprule
\# & Attribute Name & \# & Attribute Name \\
\midrule

1 & ID & 23 & DTCPB \\
2 & DUR & 24 & DWIN \\
3 & PROTO & 25 & TCPRTT \\
4 & SERVICE & 26 & SYNACK \\
5 & STATE & 27 & ACKDAT \\
6 & SPKTS & 28 & SMEAN \\
7 & DPKTS & 29 & DMEAN \\
8 & SBYTES & 30 & TRANS DEPTH \\
9 & DBYTES & 31 & RESPONSE BODY LEN \\
10 & RATE & 32 & CT SRV SRC \\
11 & STTL & 33 & CT STATE TTL \\
12 & DTTL & 34 & CT DST LTM \\
13 & SLOAD & 35 & CT SRC DPORT TM \\
14 & DLOAD & 36 & CT DST SPORT ITM \\
15 & SLOSS & 37 & CT DST SRC ITM \\
16 & DLOSS & 38 & IS FTP LOGIN \\
17 & SINPKT & 39 & CT FTP CMD \\
18 & DINPKT & 40 & CT FLW HTTP MTHD \\
19 & SJIT & 41 & CT SRC ITM \\
20 & DJIT & 42 & CT SRV DST \\
21 & SWIN & 43 & IS SM IPS PORTS \\
22 & STCPB & 44 & ATTACK CAT \\
 & & 45 & LABEL \\
\hline
\end{tabular}
\end{table}

%\begin{table}[!t]
%\centering
%\caption{UNSW-NB15 Dataset Features List}
%\label{tab1}
%
%\begin{tabbing}
%\hspace{1cm} \= \hspace{4cm} \= \hspace{1cm} \= \hspace{4cm} \= \kill % Define tab stops
%
%\# \> Attribute Name \> \# \> Attribute Name \\
%
%1 \> ID \> 23 \> DTCPB \\
%2 \> DUR \> 24 \> DWIN \\
%3 \> PROTO \> 25 \> TCPRTT \\
%4 \> SERVICE \> 26 \> SYNACK \\
%5 \> STATE \> 27 \> ACKDAT \\
%6 \> SPKTS \> 28 \> SMEAN \\
%7 \> DPKTS \> 29 \> DMEAN \\
%8 \> SBYTES \> 30 \> TRANS DEPTH \\
%9 \> DBYTES \> 31 \> RESPONSE BODY LEN \\
%10 \> RATE \> 32 \> CT SRV SRC \\
%11 \> STTL \> 33 \> CT STATE TTL \\
%12 \> DTTL \> 34 \> CT DST LTM \\
%13 \> SLOAD \> 35 \> CT SRC DPORT TM \\
%14 \> DLOAD \> 36 \> CT DST SPORT ITM \\
%15 \> SLOSS \> 37 \> CT DST SRC ITM \\
%16 \> DLOSS \> 38 \> IS FTP LOGIN \\
%17 \> SINPKT \> 39 \> CT FTP CMD \\
%18 \> DINPKT \> 40 \> CT FLW HTTP MTHD \\
%19 \> SJIT \> 41 \> CT SRC ITM \\
%20 \> DJIT \> 42 \> CT SRV DST \\
%21 \> SWIN \> 43 \> IS SM IPS PORTS \\
%22 \> STCPB \> 44 \> ATTACK CAT \\
%\> \> 45 \> LABEL \\
%\end{tabbing}
%
%\end{table}


\begin{table}[!t]
\caption{UNSW-NB15 Distribution Sample\label{tab4}}
\centering
\begin{tabular}{ l l l l l }
\hline
\textbf{Category} & \textbf{Training set Size} & \textbf{Training set Distribution} & \textbf{Testing set Size} & \textbf{Testing set Distribution} \\
\hline
Attack Packets & 119,341 & 68.06 & 45,332 & 55.06 \\
\hline
Normal Packets & 56,000 & 31.94 & 37,000 & 44.94 \\
\hline
Overall Samples & 175,341 & 100 & 82,332 & 100 \\
\hline
\end{tabular}
\end{table}

\subsection{Dataset preparation and feature encoding}\label{subsec5}
\begin{figure}[!t]%
\graphicspath{ {./pictures/} }
\centering

\includegraphics[width=0.75\textwidth]{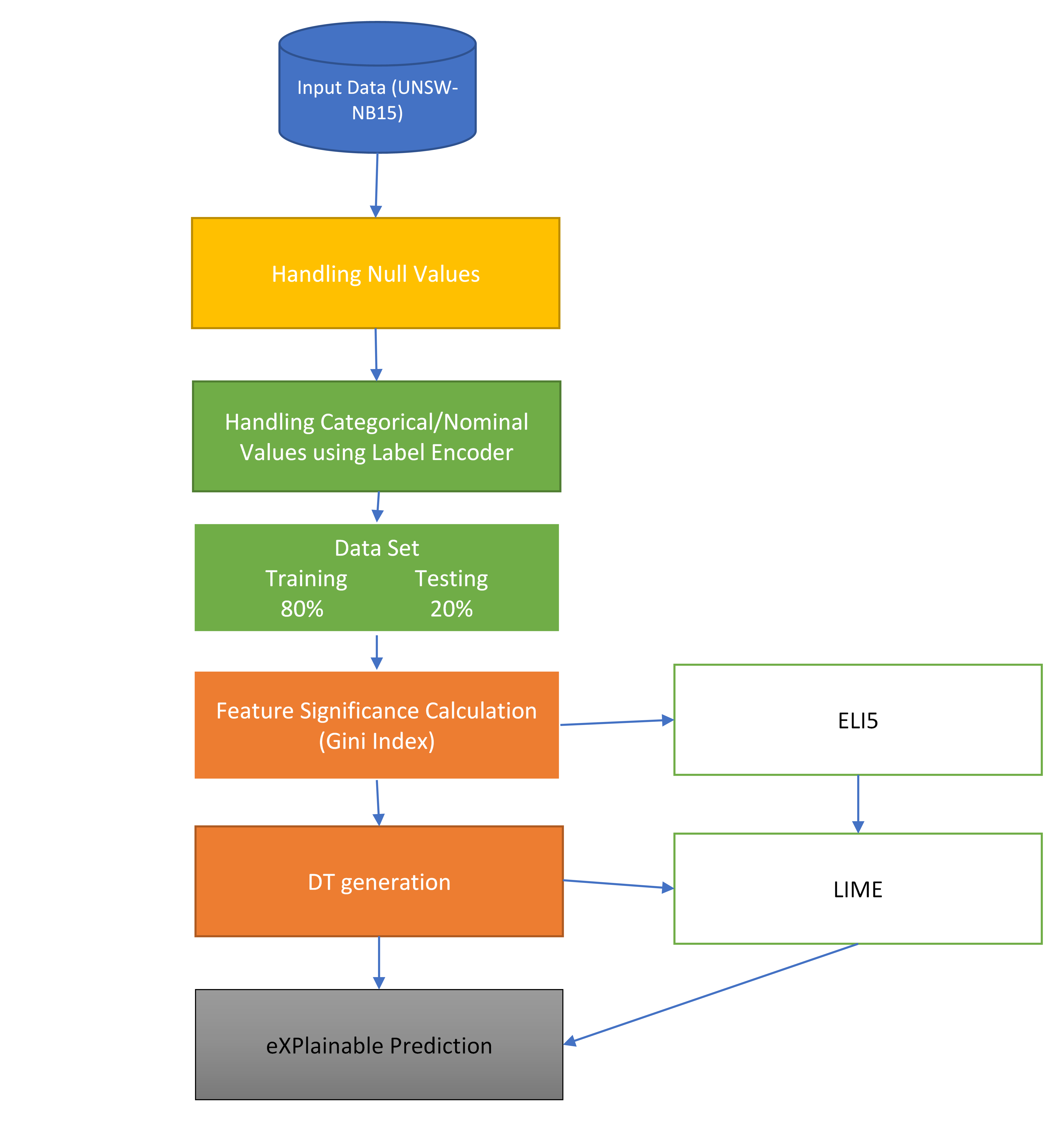}
\caption{Flow of proposed algorithm}\label{fig8}

\end{figure}

\quad First two steps of Fig. \ref{fig8} show the pre-processing steps for the dataset preparation.  Data preparation comprises both feature encoding and scaling. The first step is to handle the null values present in the dataset and converting the categorical data in numerical form with the help of label encoder. Second step is to handle categorical and nominal values via Label encoder. The third step is to split the dataset in 80-20 split for training and testing sets. Fourth step is to calculate feature significance via Gini Index and populates ELI5 for global explanation. Fifth step in the process is Classification and Regression Tree generation and populating LIME for local explanation which in combination with ELI5's global explanation provides explanation to the packet chosen for analysis 
The original UNSW-NB15 dataset features 49 attributes but we combine the four attributes from Flow category (Script, Sport, Dstip. Dsport, Proto) to make one single attribute named ID and two attributes from time category (ltime and stime) are combined to make Rate attribute. Hence, bringing the count down to 45 attributes only. 

The UNSW-NB15 dataset comprises of a mix of the numeric and categorical values. Although most of the features are valued numerically, some of the attributes are nominally valued, such as protocol type, attack category. Hence, the subsequent action involves transforming all the nominal value attributes into vectors to fit the attribute data for the IDS model. For this purpose, we used “Label Encoding” technique to directly convert the feature value to a specific numeric values. On the contrary. using One hot encoding might increase the feature dimensions. 
\subsection{Feature Ranking}\label{subsec6}
\quad The next step after data preparation and encoding is to generate a feature significance ranking of the features and select the most important ones for further processing. A feature significance ranking generally provides a numerical score and theoratical insight in how critical a feature is for building a classification and regression based tree for intrusion detection. We determine the importance of features by assessing the reduction in node impurity, factoring in the likelihood of reaching at a particular node. For measuring node impurity, we use "Gini Index" to measure the wrongly detected random element. Gini Index provides the probability of wrongful classification/categorization of a specific attribute in the dataset with respect to class distribution. 
The feature that decreases the impurity more is consequently considered the more important feature. After calculation of Gini Index for impurity for every attribute, we rank them with respect to their score in descending order and then select N most important features depending on the preference of the end-user i-e cybersecurity engineer. This facilitates the creation of an efficient decision tree and allows us to discern associations and patterns within the dataset. Consequently, we can restructure these into datasets of significantly lower dimensions without forfeiting any critical information necessary for model design and performance. 
The visualization of the attribute significance is done by using ELI5 library that assists in bringing explainability to the steps being conducted.

\subsection{Designing CART based tree}\label{subsec7}

\quad Once the security attributes are ranked in descending order, we design a classification and regression tree model that makes data-driven intrusion detection decisions. This model only accounts for the N number of most significant features according to the feature significance score calculated and the number specified by the human-in-the-loop. Rooting out the less significant attributes might have an negative effect on accuracy but also reduces the processing time hence making the mission critical system like IDS to process information on the run-time. The tree-like model initiates with a root node and breaks down the training set to smaller subsets and every associated brance is created incrementally. The Gini Index is used to to identify the root node's attribute at each nodel level. The feature with significantly lower Gini index is chosen to build the tree and consequently expanding it by generating the specified branch number comprising of internal and leaf nodes along with the connecting arcs. All the inner nodes are tagged with the earlier chosen attributes and each leaf node is tagged with a category i-e normal or attack category in this instance. 

\begin{figure*}[!t]%
\graphicspath{ {./pictures/} }
\centering
{\includegraphics[width=0.9\textwidth]{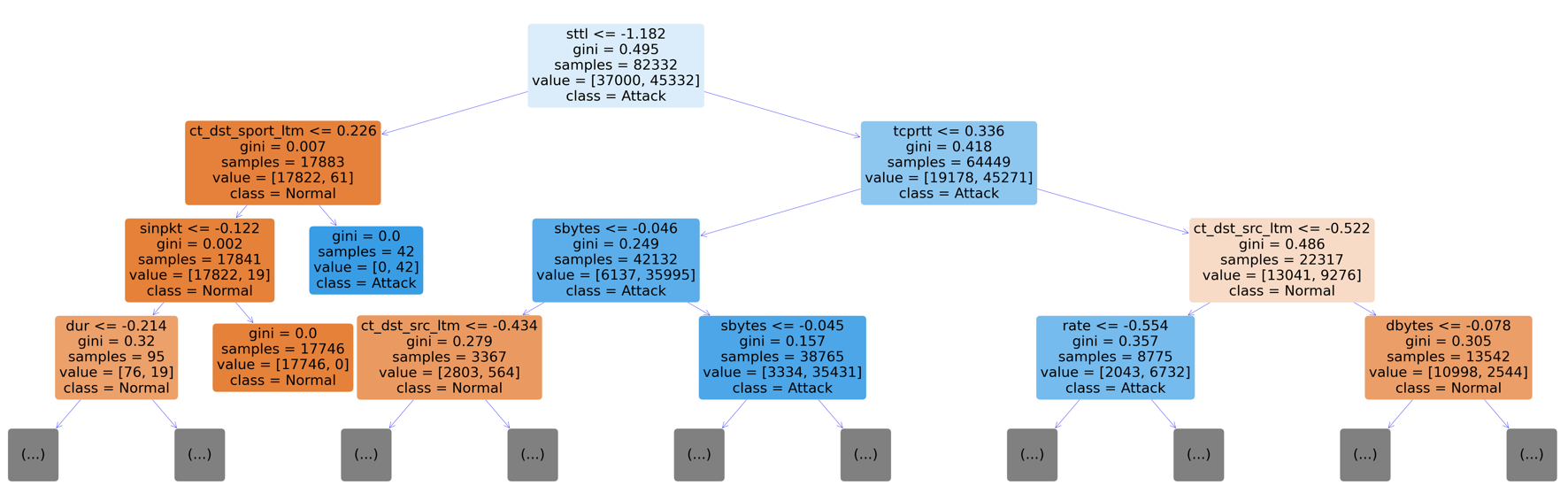}}
\caption{Multi-level tree with child nodes and Gini impurity}\label{fig9}
\end{figure*}

The end result is, we get a multi-level tree with several child nodes each specifying the attack or normal class category as shown in Figure \ref{fig9}. Our explainable IDS model in particular has two significant properties:
\begin {enumerate}[1.]
\item reduced feature dimensions due to feature significance score and ranking
\item construction of  a multi-level decision tree considering based on the N number of signifcant features as specified by the end-user (human-in-the-loop/overseer) 
\end{enumerate}
Figure \ref{fig9} illustrates a sample output of this L-XAIDS considering several features like flow features (srcip, dstip), basic features (state, duur, sbytes, dbytes), content features (swin, dwin, stcpb, dtcpb).

\section{Results}\label{sec5}
\subsection{Decision Tree Classifier}\label{subsec8}
\begin{figure}[!t]%
\graphicspath{ {./pictures/} }
\centering

\includegraphics[width=0.75\textwidth]{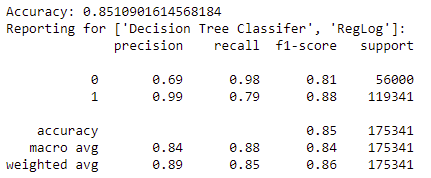}
\caption{Decision Tree Classifier results}\label{fig10}

\end{figure}

\begin{figure}[!t]%
\graphicspath{ {./pictures/} }
\centering

\includegraphics[width=0.75\textwidth]{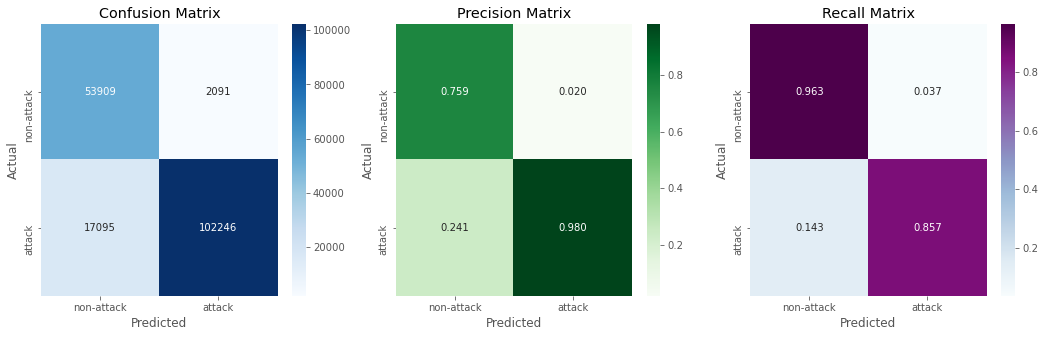}
\caption{Confusion-Precision-Recall Matrix}\label{fig11}

\end{figure}

\quad We utilized the tree.DecisionTreeClassifier() module from Scikit-learn library to develop a decision tree classification model to classify network traffic behavior as Normal or Attack. The accuracy of the model against the testing set was found to be 85 percent, as illustrated in Fig. \ref{fig10} and Fig. \ref{fig11}.

\begin{figure}[!t]%
\graphicspath{ {./pictures/} }
\centering

\includegraphics[width=0.75\textwidth]{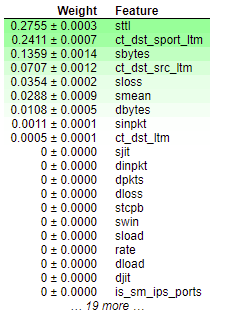}
\caption{ELI5 Permutation Importance for CART}\label{fig12}
\end{figure}

\begin{figure}[!t]%
\graphicspath{ {./pictures/} }
\centering

\includegraphics[width=0.75\textwidth]{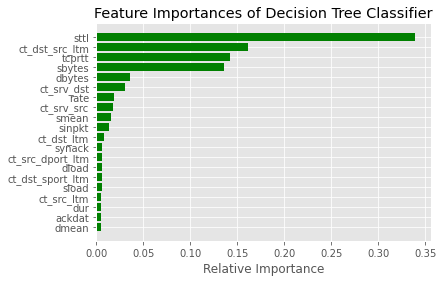}
\caption{Scikit learn feature importance for CART}\label{fig13}
\end{figure}

To determine the significance of features, we employed the ELI5 Permutation Importance toolkit along with scikit-learn to generate a graph for the top ten features. The feature importance is calculated based on the reduction in node impurity value, weighted by the probability of reaching that node. The output of feature significance from both ELI5 Permutation Importance toolkit (Fig. \ref{fig12}) and scikit-learn (Fig. \ref{fig13}) indicates that the ‘sttl’ feature, i.e., “source to destination time to live value,” is the most important attribute for classification prediction in network traffic analysis. The generated visual of Decision Trees will depict the most significant attributes in descending order, as shown in Fig. \ref{fig9}.

The evaluation metrics used in this study include precision, recall, and F1-score, which inherently capture both False Positive Rate (FPR) and True Positive Rate (TPR). Specifically, recall is equivalent to TPR, as it measures the proportion of correctly identified attack instances. Additionally, precision accounts for false positives, which indirectly reflects FPR. The F1-score provides a harmonized measure of both precision and recall, ensuring a comprehensive evaluation of IDS performance. Given these metrics, we ensure a robust assessment of classification effectiveness without redundancy.

The use of decision tree visualizations enables the explanation of models through the examination of each decision level, along with its associated feature and splitting value for each condition.  Furthermore, the classification prediction result for each class line is illustrated based on the chosen maximum depth of the tree. The application of decision trees for ML-based IDSs yields highly accurate classification outcomes, indicating effective detection of malicious threats. In addition, the DT algorithm's explainability features, as demonstrated by plotting decision trees, can aid human analysts in comprehending the model. This understanding includes the ability to speculate on what the IDS machine learned from the features or compare expectations.

In addition to model accuracy and explainability, computational efficiency is a key consideration for real-world IDS deployment. The LIME-based local explanation process took an average of 2.3 seconds per instance, while ELI5’s permutation importance computation required approximately 8.5 seconds per feature subset. While these explainability techniques introduce some computational overhead, the trade-off is justified as IDS analysts require interpretable decisions for incident response and forensic analysis. Future optimizations could leverage GPU acceleration or parallel computation to improve runtime performance.

\begin{figure*}[!t]%
\graphicspath{ {./pictures/} }
\centering
\includegraphics[width=0.9\textwidth]{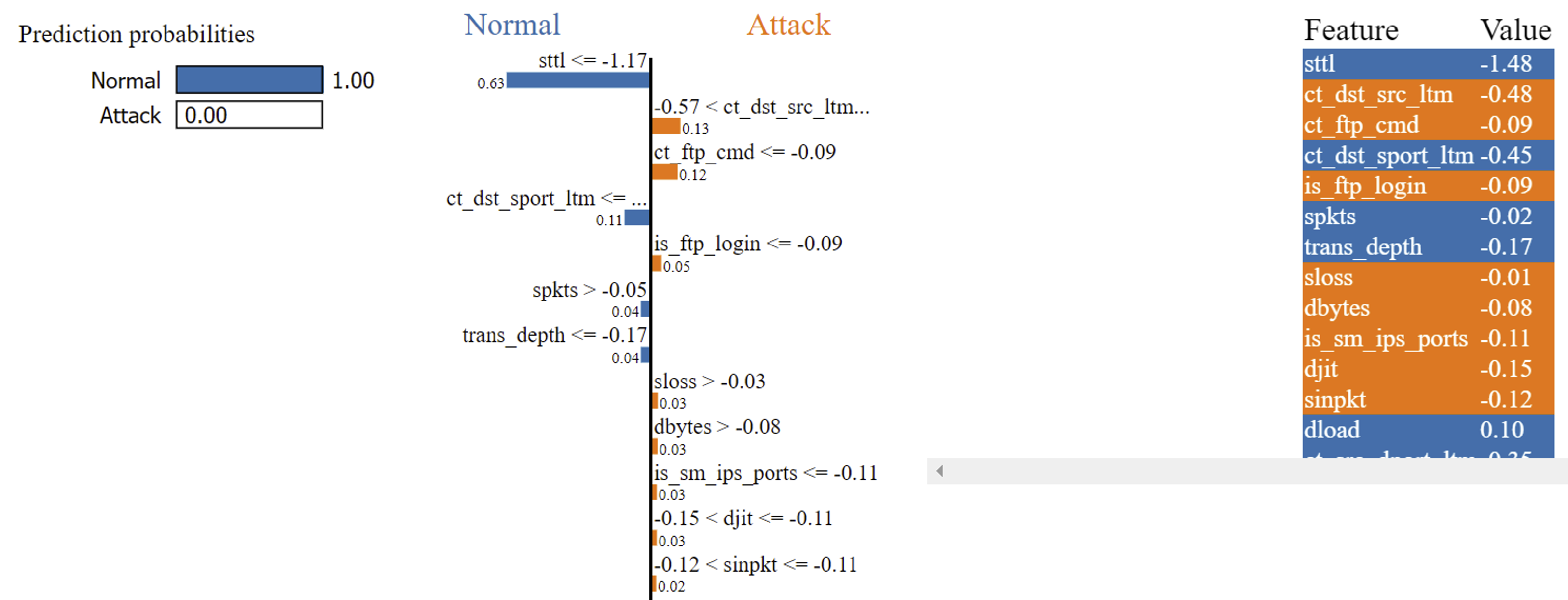}
\caption{Binary classification prediction using LIME}\label{fig14}
\end{figure*}

\subsection{Multi-layer Perceptron (MLP) Classifier}\label{subsec9}
\quad Multilayer perceptron (MLP) is a type of artificial neural network that can learn and make predictions based on a set of input features. In the context of intrusion detection, MLP classifiers have been applied to datasets such as UNSW-NB15 to classify network traffic as either normal or malicious. However, the predictions of the MLP model can be difficult to interpret, which limits its usefulness in real-world scenarios. To provide explainability, methods such as LIME, ELI5, and scikit-learn are used to highlight which input features are most influential in the model's decision-making process. LIME (Local Interpretable Model-Agnostic Explanations) and ELI5 (Explain Like I'm 5) are both Python libraries that can generate human-readable explanations of the model's predictions, while scikit-learn is a machine learning library that provides tools for model interpretation and feature selection. By applying these methods to the MLP classifier trained on UNSW-NB15 dataset, we can gain insights into which network traffic features are most important for detecting intrusions and explain how the model functions.

\begin{figure*}[!t]%
\graphicspath{ {./pictures/} }
\centering
\includegraphics[width=0.9\textwidth]{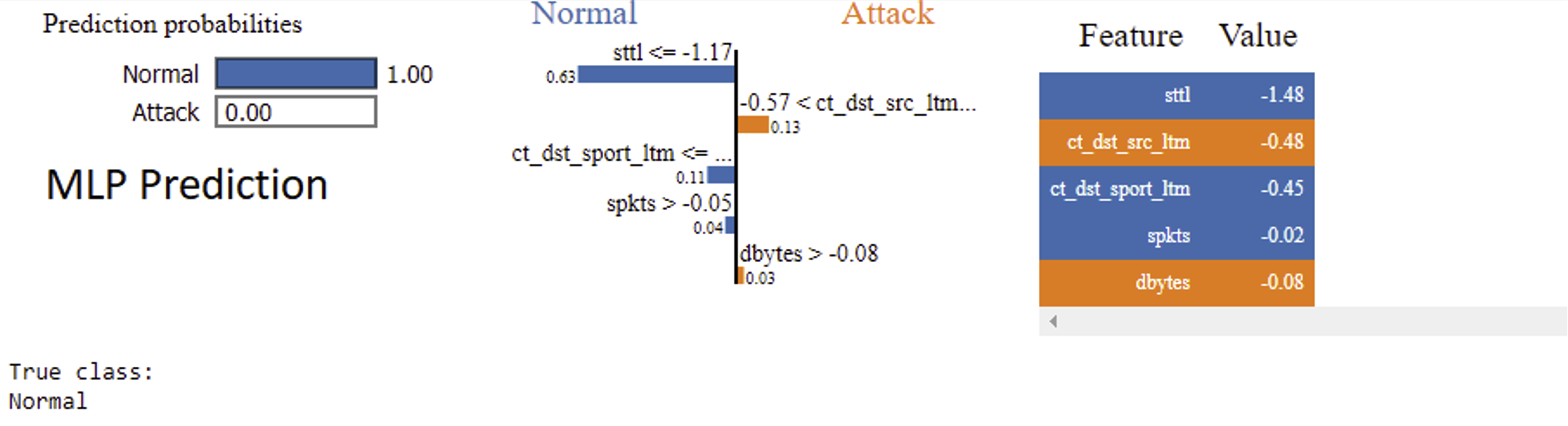}
\caption{LIME for MLP Classifier Explanation}\label{fig15}
\end{figure*}

The MLP classifier underwent training and testing on respective datasets, resulting in an outstanding model performance accuracy of 89.83 percent when tested against the test set. This remarkable score indicated that the MLP classifier was capable of detecting normal or attack behavior in network traffic with high precision. In order to generate a visual representation of the MLP Classifier's prediction for individual predictions in the training set, we utilized the LIME library. Based on the proximity of the new data outputs to the original point, the library assigns them weights and fits a surrogate linear regression on the dataset with variations using the sample weights. The explanation model that is thus created is used to explain the original data points. Figure \ref{fig15} illustrates Lime Tabular Explainer output highlighting top five features. The LIME library's use of surrogate models to explain predictions provides a deeper understanding of how the model is making its classifications. The visual representation generated using this approach can enable security analysts to gain deeper insights into how the model arrived at its conclusions and identify possible data biases that can be corrected to improve future model performance.
The visual dashboard is a useful tool that provides an explanation for the predicted behavior classification of network traffic records. It displays the features and their weights that contributed to the classification of the traffic record as "Normal". The accuracy of the classification is confirmed by comparing it to the true class label of "Normal". This dashboard provides a detailed explanation of the predicted classifications, which allows human analysts to conduct thorough investigations into cybersecurity incidents or conduct follow-up assessments to understand why certain network traffic was classified in a particular way by the model. By combining the transparency of the visual dashboard with the high-performance benefits of a neural net MLP classifier, which is commonly considered a "black box", this tool provides an increased level of transparency for predictions that can be useful for future cybersecurity research. This level of transparency can be leveraged to better understand how the model is making predictions and improve its performance.

\subsection{XGBoost Classifier}\label{subsec10}

\begin{figure*}[!t]%
\graphicspath{ {./pictures/} }
\centering
\includegraphics[width=0.9\textwidth]{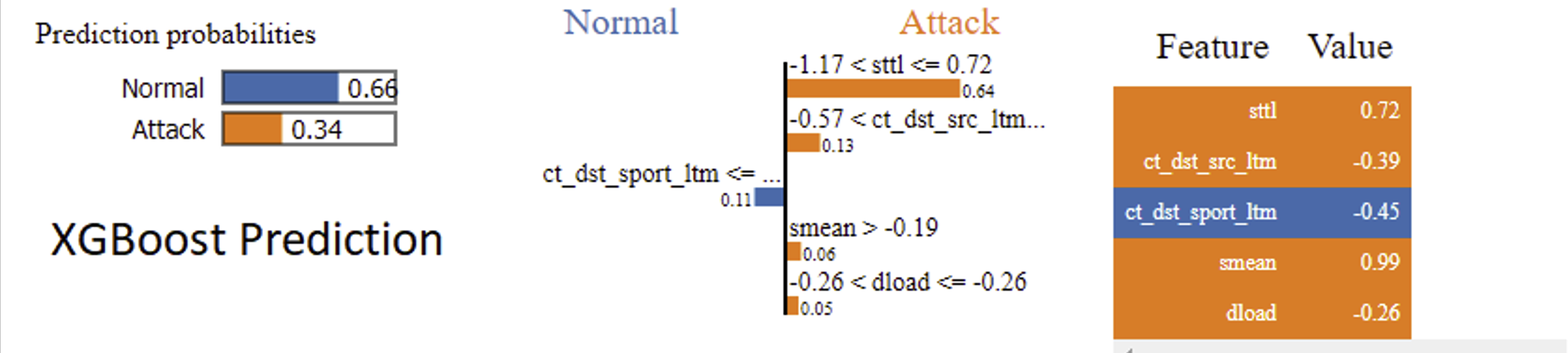}
\caption{LIME for XGBoost Classifier Explanation}\label{fig16}
\end{figure*}

\quad XGBoost is an ensemble learning algorithm that combines multiple decision trees to improve prediction accuracy. When applied to the UNSW-NB15 dataset, XGBoost can provide effective intrusion detection capabilities. To provide explainability, techniques like LIME, ELI5, or scikit-learn were utilized. For example, LIME generates explanations in the form of feature importances, showing the contribution of each feature to the final classification decision. ELI5, on the other hand, provides model-agnostic explanations, such as showing the most important features for different classes. Scikit-learn also provides insight into how the model makes decisions, by analyzing the feature importance scores generated by XGBoost. Overall, XGBoost with explainability techniques can be an effective approach for intrusion detection, providing not only accurate predictions but also insights into how the model makes decisions.
The XGBoost Classifier was also trained and tested on the testing set, and its performance was comparable to that of the MLP Classifier. To provide explainability to XGBoost Classifier, the LIME library was used to analyze the features that have the greatest impact on the output (as shown in Fig. \ref{fig16}). This classifier is highly efficient, flexible, and high-performing, and can be paired with the LIME library to provide robust explainability features.

\subsection{Significance of the explainable module}\label{subsec10}

\quad Fig. \ref{fig12} and \ref{fig13} illustrate the weights given by scikit-learn and ELI5 for global explanations that help us identifying the most significant attributes for the dataset and packet analysis. When analyzing the scikit-based weights and the permutation-based weights, we observe there is not much difference in the global identifiers.

\subsubsection{LIME Explanation for binary classifier}\label{subsubsec1}
\quad Fig. \ref{fig14} presents the LIME explanations for one of the network packets belonging to test set for UNSW-NB-15. The left side of the figure illustrates the prediction being made by the classifier itself while the right hand side shows the explanations being generated by the LIME. 
Upon comparison of these LIME visualizations with the significant global features, it is seen that the top features presented by LIME match those given by the global permutation. 

The prediction for this packet is :
\begin {itemize}
\item $Normal: 1.00$ probability
\item $Attack/Malicious: 0$ probability
\end {itemize}
The most heavily weighted variable according to LIME is $"sttl"$ (source-to-destination time to live value) and it indicates normal behaviour for this packet. The other highly weighted attributes are $"CT\_DST\_SRC\_ITM"$ (No. of connections of the same source and the destination address in 100 connections), $"CT\_FTP\_CMD"$ (No. of flows that has a command a in ftp session), $"CT\_DST\_SPORT\_ITM"$ (Number of connections of the same destination address and the source port in 100 connections), $"IS\_FTP\_LOGIN"$ (if the ftp session is accessed by the user and password). 

Likewise, the overall LIME feature weights on the right side are:

\begin {itemize}
\item $Normal: 0.63+0.11+0.04+(0.02x4)=0.86$

\item $Attack: 0.13+0.12+0.05+(0.03x4)+(0.02x4)+0.01=0.51$
\end {itemize}
The lime explanations align with the prediction made and list almost all the significant attributes as listed by global explanations. 

\subsubsection{LIME Explanation for MLP}\label{subsubsec2}
\quad Fig. \ref{fig15} displays the LIME explanations for the tenth packet in the test set for UNSW-NB-15. The left hand side of the figure shows the prediction being made by the MLP classifier while the right hand side shows the explanations being generated by the LIME for the MLP classifer prediction. 

Similar to the datapoint $1$, the LIME visualizations/explanations enforce the global features importance generated. 
The prediction for this packet is :
\begin {itemize}
\item $Normal: 1.00$ probability
\item $Attack/Malicious: 0$ probability
\end {itemize}

The $"sttl"$ attribute appears to be the most significant feature according to LIME, pointing towards Normal, followed by $"CT\_DST\_SRC\_ITM"$ (No. of connections of the same source and the destination address in 100 connections), $"CT\_DST\_SPORT\_ITM"$ (Number of connections of the same destination address and the source port in 100 connections), $"SPKTS"$ (Source to destination packet count), and $"DBYTES"$ (Destination to source transaction bytes).

Likewise, the LIME feature weights on the right side are as follows:
\begin {itemize}
\item $Normal: 0.63+0.11+0.04=0.78$
\item $Attack: 0.13+0.03=0.16$
\end {itemize}
The lime explanations align with the prediction made and list almost all the significant attributes as listed by global explanations. 

\subsubsection{LIME Explanation for XGBoost}\label{subsubsec3}
\quad Fig. \ref{fig16} shows the LIME explanations for the hundredth packet in the test set for UNSW-NB-15. The left side of the picture shows the prediction being made the XGBoost classifier itself while the right hand side shows the explanations being generated by the LIME. 

Now, if we collate these LIME visualizations/explanations for with the global feature importances above , we observe that the top features presented by LIME closely align with those given by the global permutation importance. 
The prediction for this packet is :
\begin {itemize}
\item $Normal: 0.66$ probability
\item $Attack/Malicious: 0.34$ probability
\end {itemize}
According to the LIME explanation, the strongest variable affecting this prediction is $"sttl,"$ which is pointing towards "attack." Other significant attributes include $"CT\_DST\_SRC\_ITM"$ (number of connections between the same source and destination address in 100 connections), $"CT\_DST\_SPORT \_ITM"$ (number of connections between the same destination address and source port in 100 connections), $"SMEAN"$ (mean of the flow packet bytes transmitted by the source), and $"DLOAD"$ (destination bits per second).

The LIME feature weights on the right side of the figure show a much stronger emphasis towards "attack" with a total weight of $0.88$, while the actual classifier is leaning towards "normal," with a weight of only $0.11$.
\begin {itemize}
\item $Normal: 0.11$
\item $Attack: 0.64+0.13+0.06+0.05=0.88$
\end {itemize}

This is a case of false negative, which could potentially harm the security of a corporate environment and result in service or financial losses. This test point shows the application and significance of local and global explanations where LIME along with ELI5 and scikit-learn can be used to make sense of predictions made and provide better insight into working of the algorithm. 

\subsubsection{Summary of Local/Global Explanation Results}\label{subsubsec11}
This section delves into the local and global interpretations of L-XAIDS. Figures \ref{fig14} to \ref{fig16} exhibit the contribution level of each feature value to the decisions made by the models. Each feature value effects the accumulated prediction value, either enhancing or diminishing the prediction.

Figure \ref{fig14} depicts the top ten influential features.: 
\begin {itemize}
\item sttl
\item ct\_cst\_src\_ltm
\item ct\_ftp\_cmd
\item ct\_dst\_sport\_ltm
\item is\_ftp\_login
\item spkts
\item trans\_depth
\item sloss
\item dbytes
\item is\_sm\_ips\_ports
\end {itemize}

And these significant atrributes' values  will contribute in determining the data as normal or malicious traffic  Therefore, the weights assigned along with lime visual predictions provide solid pieces of evidence to cybersecurity experts for auditing of the judgments made by the IDS's.

Fig. \ref{fig12} and \ref{fig14} show the top 10 features extracted for attack and normal traffic in binary classifier. These features drive the dataset trend and provide global explanations on the significance of each attribute for binary classification. The more significant features are illustrated from top to bottom in a darker shade of green for ELI5 (Fig. \ref{fig12}) and by bar values for scikit-learn (Fig.\ref{fig13}).  The values of the features increase as the green color intensity grows. A summary of the differences between the top 10 crucial features obtained by ELI5 and scikit-learn is presented below:
\begin {itemize}
\item 9 out of 10 features are identical for global explanations for both ELI5 and scikit-learn.
\item Similar to local explanations sttl, ct cst src ltm and cs dst sport ltm are the most significant attributes for global explanation.
\item There is also some differences in the order of the significance of the attributes.
\end {itemize}

While sttl (source to destination time-to-live) emerged as the most significant feature in distinguishing between normal and attack traffic, further validation was conducted to assess its role in misclassification errors. Analysis of false positives and false negatives revealed that while sttl plays a critical role in attack detection, certain traffic patterns—especially those with varying time-to-live (TTL) values—resulted in misclassified instances. These errors were more prevalent in attack samples where TTL values closely resembled normal traffic, leading to false negatives. This finding highlights the necessity of considering multiple feature interactions, and our framework enables security analysts to interpret feature contributions more effectively, reducing misclassification risks in IDS decision-making.

\subsubsection{Applications of the proposed algorithm}\label{subsubsec4}

\quad The proposed technique to combine global and local explanations will enhance SOC teams ability to flag the otherwise false negative predictions. These false negative predictions might result in severe impact to business critical systems, and the transparency provided by the explainable module of IDS will assist in detecting wrongly labelled traffic in real-time and, consequently assist in improving the configurations and auditability of the systems deployed. Explainable IDS is an efficient way to visualize feature importances for a datapoint.

This hybrid explainable technique assists in identifying the important features for malware families. The primary focus of this proposed framework is to improve the explainability of the IDS. Hence, as an add-on to classifier predictions, local and global visual and numerical explanations are also provided for SOC teams to improve their trust in the IDS decision making and to bring transparency to the process. This framework can ultimately assist security professionals to develop a better undestanding of IDS internal functions. 

\subsubsection{Comparative Analysis of XAI Frameworks: ML Algorithms, Explainability Techniques, and Scope}\label{subsubsec5}
Analyzing explainability in terms of IDS (Intrusion Detection System) algorithms involves evaluating how well the algorithms can provide insights into their decision-making process. This can be done through various metrics such as interpretability, transparency, and fidelity. Interpretability refers to the ability of the IDS algorithm to provide clear and understandable explanations of its decision-making process, while transparency measures the extent to which the algorithm can provide a clear view of the internal workings. Fidelity, on the other hand, measures the accuracy of the explanations provided by the algorithm in reflecting the actual decision-making process. 
Here is a comparison of ten research studies on machine learning-based IDS algorithms conducted since 2021 with respect to their explainability approach, type of explainability (local, global, or both), dataset used, algorithms used for both machine learning and explainability, name of algorithms, and level of explainability, along with citations of research papers and authors

\begin{table*}[ht!]
\centering
\caption{Comparison of explainable IDS frameworks\label{tab3}}
\begin{tabular}{|p{1.5cm}|p{3cm}|p{3cm}|p{1.5cm}|p{0.5cm}|p{0.5cm}|p{0.5cm}|}
\hline
\textbf{Author} & \textbf{ML Algorithm} & \textbf{Explainability Algorithm} & \textbf{Scope of Explainability} & \textbf{E1} & \textbf{E2} & \textbf{E3} \\
\hline

\cite{R55} & Hybrid Sampling & Grad-CAM & Global & \checkmark & \ding{55} & \ding{55} \\
\hline
\cite{R56} & Light GBM & SHAP & Global & \checkmark & \ding{55} & \checkmark \\
\hline
\cite{R57} & Random Forests & SHAP & Global & \checkmark & \ding{55} & \checkmark \\
\hline
\cite{R58} & SVM & LIME & Local & \ding{55} & \checkmark & \checkmark \\
\hline
\cite{R59} & Balanced Stacked RF & SHAP & Global & \checkmark & \ding{55} & \checkmark \\
\hline
\cite{R60} & Attention Mechanism & Attention Map & Global & \checkmark & \ding{55} & \ding{55} \\
\hline
\cite{R61} & CNN & Feature and Model Exploration & Local & \checkmark & \checkmark & \ding{55} \\
\hline
\cite{R62} & Hybrid of ML/DL & Performance Analysis & Global & \checkmark & \checkmark & \ding{55} \\
\hline
L-XAIDS* & Decision Tree, MLP, XGBoost & LIME/ELI5/Scikit-learn & Local, Global & \checkmark & \checkmark & \checkmark \\
\hline

\end{tabular}
\end{table*}

To conduct comparative analysis for evaluating the explainability provided by similar research works, we use the Degree of explainability (DoX) explanandum aspects as proposed in \cite{R76}
Specifically, we consider the following explanandum aspects  $E$ for each B2B requirements :
\begin{itemize}
\item Explicitly highlighting the main features used in decision making by the AI: where $E1$ is the set of main attributes  used for decision making.
\item Providing a list of all attributes processed by the AI: where $E2$ is the set of all the attributes being processed by the
AI.
\item Providing an explanation of a specific decision taken by the AI: where $E3$ is the decision based on the attributes being evaluated and coorelate to the explanation provided. \cite{R77}.
\end{itemize}
Table 3 presents a summary of comparison of similar work with the proposed framework based on these explanandums.

L-XAIDS, our proposed explainable IDS algorithm, combines LIME, ELI5, and Scikit-learn to provide local and global explanations for decision tree, MLP, and XGBoost models trained on the UNSW-NB15 dataset. Compared to other similar algorithms, our approach has several advantages.
\begin {itemize}
\item First, L-XAIDS provides both local and global explanations for the classifiers, which enables users to understand why certain decisions were made at the instance-level as well as the overall behavior of the classifier. This capability is especially important in an IDS context where the stakes are high, and the ability to interpret the decision-making process can make the difference between detecting an attack and missing it.

\item Second, L-XAIDS is flexible and can be applied to a range of classifiers, including decision trees, MLPs, and XGBoost. This versatility means that the algorithm can be used in a variety of settings and applications.

\item Third, L-XAIDS is easy to use and implement, as it is built on top of widely used machine learning libraries such as Scikit-learn, LIME, and ELI5. This makes it accessible to a broad audience, including practitioners and researchers in the IDS community.
\end {itemize}

\begin{table}[h]
    \centering
    \caption{Summarized Classification Performance Across Datasets}
    \begin{tabular}{lccc}
        \hline
        \textbf{Metric} & \textbf{UNSW-NB15} & \textbf{CSE-CIC-IDS2018} & \textbf{LITNET-2020} \\
        \hline
        \multicolumn{4}{c}{\textbf{Precision Metrics}} \\ \midrule
        Avg Precision & 0.835 & 0.805 & 0.775 \\
        Macro Avg Precision & 0.833 & 0.800 & 0.783 \\
        Weighted Avg Precision & 0.880 & 0.863 & 0.827 \\ 
        \hline
        \multicolumn{4}{c}{\textbf{Recall Metrics}} \\ \midrule
        Avg Recall & 0.883 & 0.852 & 0.815 \\
        Macro Avg Recall & 0.873 & 0.840 & 0.827 \\
        Weighted Avg Recall & 0.843 & 0.823 & 0.800 \\
        \hline
        \multicolumn{4}{c}{\textbf{F1-score Metrics}} \\ \midrule
        Avg F1-score & 0.842 & 0.820 & 0.785 \\
        Macro Avg F1-score & 0.837 & 0.803 & 0.767 \\
        Weighted Avg F1-score & 0.857 & 0.823 & 0.810 \\
        \hline
        \textbf{Overall Accuracy} & 0.85 & 0.83 & 0.81 \\
        \hline
    \end{tabular}
    \label{tab:classification_performance_grouped}
\end{table}

\begin{table}[h]
    \centering
    \caption{Additional IDS Metrics Across Datasets}
    \begin{tabular}{lccccc}
        \hline
        \textbf{Dataset} & \textbf{FPR} & \textbf{TNR} & \textbf{MCC} & \textbf{FAR} & \textbf{Balanced Accuracy} \\
        \hline
        UNSW-NB15      & 0.17  & 0.84  & 0.86  & 0.12  & 0.83  \\
        CSE-CIC-IDS2018 & 0.18  & 0.82  & 0.82  & 0.13  & 0.79  \\
        LITNET-2020    & 0.19  & 0.77  & 0.80  & 0.14  & 0.76  \\
        \hline
    \end{tabular}
    \label{tab:ids_metrics}
\end{table}

Table~\ref{tab:classification_performance_grouped} presents a comprehensive evaluation of the L-XAIDS framework’s classification performance across three different datasets: UNSW-NB15, CSE-CIC-IDS2018, and LITNET-2020. The results demonstrate that while UNSW-NB15 achieves the highest performance, the model generalizes well to other datasets with only a slight decrease in performance.

The precision metrics indicate how many of the detected attack instances were truly attacks. The high weighted average precision across all datasets confirms that false alarms are minimized, ensuring that security analysts are not overwhelmed with unnecessary alerts. The recall metrics measure how many actual attacks were correctly identified. Our model maintains high recall across datasets, proving its ability to effectively detect cyber threats in different network environments. The F1-score provides a balance between precision and recall, ensuring that neither false positives nor false negatives dominate the classification performance. Lastly, the overall accuracy, while slightly lower for CSE-CIC-IDS2018 and LITNET-2020, remains comparable, reinforcing the robustness and adaptability of L-XAIDS across different traffic patterns and attack distributions.

These findings validate that our model is not dataset-dependent and can be extended to multiple real-world cybersecurity environments without significant degradation in explainability or detection performance. L-XAIDS generalizes well across different network environments while maintaining high detection capabilities.

Table~\ref{tab:ids_metrics} evaluates the Intrusion Detection System (IDS) performance of L-XAIDS using key cybersecurity-specific metrics, which further reinforce the model’s practical effectiveness.

False Positive Rate (FPR) measures how often normal traffic is incorrectly classified as an attack. Lower FPR values indicate fewer false alarms, which is crucial for real-world IDS deployment. True Negative Rate (TNR) represents the percentage of correctly classified normal traffic. A high TNR shows that our model effectively filters out normal activity, reducing the risk of mistakenly blocking legitimate traffic. 

Matthews Correlation Coefficient (MCC) is a balanced metric that considers both false positives and false negatives. MCC values closer to 1 indicate a strong correlation between predictions and actual classifications, proving that L-XAIDS is reliable across datasets. False Alarm Rate (FAR) shows the proportion of false alerts generated by the IDS. A lower FAR ensures that our explainability-driven framework does not introduce unnecessary warnings, keeping it practical for cybersecurity analysts. Balanced Accuracy accounts for class imbalances, making it a more reliable measure in intrusion detection. L-XAIDS maintains a strong balanced accuracy, confirming that the model does not favor one class over another, making it effective even in datasets with varying attack distributions.

The results demonstrate that L-XAIDS successfully balances explainability with detection performance. By integrating LIME and ELI5 explainability techniques, the model remains interpretable without sacrificing accuracy, making it a practical solution for real-world cybersecurity applications. The ability to maintain low false alarms, high true negatives, and strong correlation metrics (MCC) solidifies L-XAIDS as a relevant and adaptable framework for mission-critical IDS deployments.

Overall, L-XAIDS provides a comprehensive and flexible approach to IDS explainability that is suitable for a wide range of classifiers and datasets. Its ability to provide both global and local explanations, combined with its ease of use and implementation, makes it a valuable tool for the IDS community.

Our framework aims to provide an explainable IDS by integrating various machine learning algorithms such as XGBoost, decision tree, and MLP classifier with LIME, ELI5, and scikit-learn. To quantify the explainability of our framework, we utilize several metrics such as the local fidelity and the consistency of the explanations. We also measure the comprehensibility of the generated explanations by conducting user studies.

In summary, our framework offers a comprehensive and effective approach for developing explainable IDS systems that can provide insights into the reasoning of the underlying models. By utilizing various machine learning algorithms along with LIME, ELI5, and scikit-learn, our framework can achieve high accuracy and interpretability while providing a high degree of user comprehensibility.

\section{Conclusion}\label{sec6}
\quad As ML models used for IDSs in network traffic security become increasingly complex, the role of human analysts with domain knowledge remains critical for resource allocation and cybersecurity strategy development. However, ML algorithms are often considered "black boxes" with outputs that lack interpretability or explanation. To address this, the UNSW-NB15 dataset was used to train three classifiers - Decision Tree, MLP, and XGBoost - and achieve high-performance overall accuracy of 85 percent in analyzing Attack or Normal activity in mission-critical networks, along with the explainability module. Subsequently, recognized libraries and methods for fostering explainability were utilized with the trained classifiers to add transparency to their decision-making processes and assess the significance of different features.. This increased transparency increases trust in ML systems in the short term and enables new capabilities in cybersecurity through extracting insights from sophisticated machine learning models. Ultimately, this will help security professionals to understand the functioning of the systems and design and configure them accordingly.

In this research project, we proposed an explainable IDS framework titled L-XAIDS that uses LIME, ELI5, and scikit-learn for providing local and global explanations for decision tree, MLP classifier, and XGBoost on the UNSW-NB15 dataset. Our framework offers increased transparency and explainability capabilities for complex ML models in network traffic security. Our results showed that the L-XAIDS framework achieved high-performance accuracy in analyzing Attack or Normal activity in mission-critical networks. Furthermore, the explainability techniques used in this framework provided insights into the decision-making process of the classifiers and feature importance evaluation.

This study introduced L-XAIDS, a novel explainable AI framework that enhances the interpretability of IDS decisions. Our approach addresses the black-box nature of AI-driven cybersecurity models, ensuring transparency for security professionals. By integrating both local (LIME) and global (ELI5) explanation techniques, our framework provides meaningful insights into the decision-making process of IDSs. These explanations not only help detect attacks but also enable cybersecurity teams to audit and refine IDS models effectively.

Furthermore, the significance of explainability in AI-driven cybersecurity cannot be overstated. As AI continues to play a crucial role in security applications, the ability to understand and trust AI decisions becomes imperative. Our framework aligns with this need by offering a transparent, interpretable, and user-friendly IDS solution.

One of the key challenges in explainable AI is the trade-off between interpretability and computational efficiency. While LIME and ELI5 offer valuable insights into IDS decision-making, their runtime cost scales with dataset size. However, their integration enables cybersecurity analysts to justify automated threat detection, reducing response time in critical security operations. Future optimizations will explore hybrid explainability models that balance interpretability with minimal processing overhead.

Future research should explore how explainable AI can be integrated into real-time intrusion detection workflows and extended to other cybersecurity domains. Additionally, given the rising concerns about AI security risks, incorporating adversarial robustness into explainability techniques could enhance IDS resilience against emerging cyber threats.
Future works will focus on expanding and scaling the explainability to a decentralized explainable IDS framework and on different datasets such as CICIDS2017 and CICIDS2018, as this will offer an opportunity for stakeholders and organizations to have a deeper understanding of the functioning of the IDSs. Moreover, this will enable security professionals to make more informed decisions in real-time and address security threats promptly. The decentralized explainable IDS framework can be designed using a federated learning approach that enables the sharing of knowledge and data across different nodes in the network, while maintaining privacy and data security. Ultimately, the L-XAIDS framework and its decentralized version can assist in designing more robust, trustworthy, and effective IDSs for network security.

\section{Acknowledgments}
The authors thank the anonymous reviewers for their valuable suggestions. This work was supported by Mitacs and the partner organization Ericsson Canada Inc. through the Mitacs Accelerate program IT16752.

\bibliography{references}% common bib file

%% BioMed_Central_Bib_Style_v1.01

\begin{thebibliography}{88}
% BibTex style file: bmc-mathphys.bst (version 2.1), 2014-07-24
\ifx \bisbn   \undefined \def \bisbn  #1{ISBN #1}\fi
\ifx \binits  \undefined \def \binits#1{#1}\fi
\ifx \bauthor  \undefined \def \bauthor#1{#1}\fi
\ifx \batitle  \undefined \def \batitle#1{#1}\fi
\ifx \bjtitle  \undefined \def \bjtitle#1{#1}\fi
\ifx \bvolume  \undefined \def \bvolume#1{\textbf{#1}}\fi
\ifx \byear  \undefined \def \byear#1{#1}\fi
\ifx \bissue  \undefined \def \bissue#1{#1}\fi
\ifx \bfpage  \undefined \def \bfpage#1{#1}\fi
\ifx \blpage  \undefined \def \blpage #1{#1}\fi
\ifx \burl  \undefined \def \burl#1{\textsf{#1}}\fi
\ifx \doiurl  \undefined \def \doiurl#1{\url{https://doi.org/#1}}\fi
\ifx \betal  \undefined \def \betal{\textit{et al.}}\fi
\ifx \binstitute  \undefined \def \binstitute#1{#1}\fi
\ifx \binstitutionaled  \undefined \def \binstitutionaled#1{#1}\fi
\ifx \bctitle  \undefined \def \bctitle#1{#1}\fi
\ifx \beditor  \undefined \def \beditor#1{#1}\fi
\ifx \bpublisher  \undefined \def \bpublisher#1{#1}\fi
\ifx \bbtitle  \undefined \def \bbtitle#1{#1}\fi
\ifx \bedition  \undefined \def \bedition#1{#1}\fi
\ifx \bseriesno  \undefined \def \bseriesno#1{#1}\fi
\ifx \blocation  \undefined \def \blocation#1{#1}\fi
\ifx \bsertitle  \undefined \def \bsertitle#1{#1}\fi
\ifx \bsnm \undefined \def \bsnm#1{#1}\fi
\ifx \bsuffix \undefined \def \bsuffix#1{#1}\fi
\ifx \bparticle \undefined \def \bparticle#1{#1}\fi
\ifx \barticle \undefined \def \barticle#1{#1}\fi
\bibcommenthead
\ifx \bconfdate \undefined \def \bconfdate #1{#1}\fi
\ifx \botherref \undefined \def \botherref #1{#1}\fi
\ifx \url \undefined \def \url#1{\textsf{#1}}\fi
\ifx \bchapter \undefined \def \bchapter#1{#1}\fi
\ifx \bbook \undefined \def \bbook#1{#1}\fi
\ifx \bcomment \undefined \def \bcomment#1{#1}\fi
\ifx \oauthor \undefined \def \oauthor#1{#1}\fi
\ifx \citeauthoryear \undefined \def \citeauthoryear#1{#1}\fi
\ifx \endbibitem  \undefined \def \endbibitem {}\fi
\ifx \bconflocation  \undefined \def \bconflocation#1{#1}\fi
\ifx \arxivurl  \undefined \def \arxivurl#1{\textsf{#1}}\fi
\csname PreBibitemsHook\endcsname

%%% 1
\bibitem[\protect\citeauthoryear{Tufail et~al.}{2021}]{C1}
\begin{botherref}
\oauthor{\bsnm{Tufail}, \binits{S.}},
\oauthor{\bsnm{Parvez}, \binits{I.}},
\oauthor{\bsnm{Batool}, \binits{S.}},
\oauthor{\bsnm{Sarwat}, \binits{A.}}:
A survey on cybersecurity challenges, detection, and mitigation techniques for the smart grid.
Energies
\textbf{14}(18)
(2021)
\doiurl{10.3390/en14185894}
\end{botherref}
\endbibitem

%%% 2
\bibitem[\protect\citeauthoryear{Capuano et~al.}{2022}]{C2}
\begin{barticle}
\bauthor{\bsnm{Capuano}, \binits{N.}},
\bauthor{\bsnm{Fenza}, \binits{G.}},
\bauthor{\bsnm{Loia}, \binits{V.}},
\bauthor{\bsnm{Stanzione}, \binits{C.}}:
\batitle{Explainable artificial intelligence in cybersecurity: A survey}.
\bjtitle{IEEE Access}
\bvolume{10},
\bfpage{93575}--\blpage{93600}
(\byear{2022})
\end{barticle}
\endbibitem

%%% 3
\bibitem[\protect\citeauthoryear{Hall et~al.}{2022}]{C3}
\begin{barticle}
\bauthor{\bsnm{Hall}, \binits{S.W.}},
\bauthor{\bsnm{Sakzad}, \binits{A.}},
\bauthor{\bsnm{Choo}, \binits{K.K.R.}}:
\batitle{Explainable artificial intelligence for digital forensics}.
\bjtitle{Wiley Interdisciplinary Reviews: Forensic Science}
\bvolume{4}(\bissue{2}),
\bfpage{1434}
(\byear{2022})
\end{barticle}
\endbibitem

%%% 4
\bibitem[\protect\citeauthoryear{Dean et~al.}{2012}]{R1}
\begin{bchapter}
\bauthor{\bsnm{Dean}, \binits{J.}},
\bauthor{\bsnm{Corrado}, \binits{G.}},
\bauthor{\bsnm{Monga}, \binits{R.}},
\bauthor{\bsnm{Chen}, \binits{K.}},
\bauthor{\bsnm{Devin}, \binits{M.}},
\bauthor{\bsnm{Mao}, \binits{M.}}:
\bctitle{Large scale distributed deep networks}.
In: \bbtitle{Neural Information Processing Systems}
(\byear{2012})
\end{bchapter}
\endbibitem

%%% 5
\bibitem[\protect\citeauthoryear{Jouppi et~al.}{2017}]{R2}
\begin{bchapter}
\bauthor{\bsnm{Jouppi}, \binits{N.P.}},
\bauthor{\bsnm{Young}, \binits{C.}},
\bauthor{\bsnm{Patil}, \binits{N.}},
\bauthor{\bsnm{Patterson}, \binits{D.}},
\bauthor{\bsnm{Agrawal}, \binits{G.}},
\bauthor{\bsnm{Bajwa}, \binits{R.}}:
\bctitle{In-datacenter performance analysis of a tensor processing unit}.
In: \bbtitle{Proceedings of the 44th Annual International Symposium on Computer Architecture (ISCA)},
pp. \bfpage{1}--\blpage{12}
(\byear{2017}).
\doiurl{10.1145/3079856.3080246}
\end{bchapter}
\endbibitem

%%% 6
\bibitem[\protect\citeauthoryear{Raina et~al.}{2009}]{R3}
\begin{bchapter}
\bauthor{\bsnm{Raina}, \binits{R.}},
\bauthor{\bsnm{Madhavan}, \binits{A.}},
\bauthor{\bsnm{Ng}, \binits{A.Y.}}:
\bctitle{Large-scale deep unsupervised learning using graphics processors}.
In: \bbtitle{Proceedings of the 26th Annual International Conference on Machine Learning (ICML)},
\bconflocation{New York, NY, USA},
pp. \bfpage{873}--\blpage{880}
(\byear{2009}).
\doiurl{10.1145/1553374.1553486}
\end{bchapter}
\endbibitem

%%% 7
\bibitem[\protect\citeauthoryear{Dean and Ghemawat}{2004}]{R4}
\begin{bchapter}
\bauthor{\bsnm{Dean}, \binits{J.}},
\bauthor{\bsnm{Ghemawat}, \binits{S.}}:
\bctitle{Mapreduce: Simplified data processing on large clusters}.
In: \bbtitle{Proceedings of the 6th Conference on Symposium on Operating Systems Design and Implementation - Volume 6 (OSDI)}
(\byear{2004})
\end{bchapter}
\endbibitem

%%% 8
\bibitem[\protect\citeauthoryear{Gonzalez et~al.}{2012}]{R5}
\begin{bchapter}
\bauthor{\bsnm{Gonzalez}, \binits{J.E.}},
\bauthor{\bsnm{Low}, \binits{Y.}},
\bauthor{\bsnm{Gu}, \binits{H.}},
\bauthor{\bsnm{Bickson}, \binits{D.}},
\bauthor{\bsnm{Guestrin}, \binits{C.}}:
\bctitle{Powergraph: Distributed graph-parallel computation on natural graphs}.
In: \bbtitle{Operating Systems Design and Implementation (OSDI)},
pp. \bfpage{17}--\blpage{30}
(\byear{2012})
\end{bchapter}
\endbibitem

%%% 9
\bibitem[\protect\citeauthoryear{Zaharia et~al.}{2012}]{R6}
\begin{bchapter}
\bauthor{\bsnm{Zaharia}, \binits{M.}},
\bauthor{\bsnm{Chowdhury}, \binits{M.}},
\bauthor{\bsnm{Das}, \binits{T.}},
\bauthor{\bsnm{Dave}, \binits{A.}},
\bauthor{\bsnm{Ma}, \binits{J.}}:
\bctitle{Resilient distributed datasets: A fault-tolerant abstraction for in-memory cluster computing}.
In: \bbtitle{Networked Systems Design and Implementation (NSDI)}
(\byear{2012})
\end{bchapter}
\endbibitem

%%% 10
\bibitem[\protect\citeauthoryear{Bergstra et~al.}{2010}]{R7}
\begin{bchapter}
\bauthor{\bsnm{Bergstra}, \binits{J.}},
\bauthor{\bsnm{Breuleux}, \binits{O.}},
\bauthor{\bsnm{Bastien}, \binits{F.}},
\bauthor{\bsnm{Lamblin}, \binits{P.}},
\bauthor{\bsnm{Pascanu}, \binits{R.}},
\bauthor{\bsnm{Desjardins}, \binits{G.}}:
\bctitle{Theano: a cpu and gpu math expression compiler}.
In: \bbtitle{Proceedings of the Python for Scientific Computing Conference (SciPy)},
vol. \bseriesno{4}.
\bconflocation{Austin, TX},
p. \bfpage{3}
(\byear{2010})
\end{bchapter}
\endbibitem

%%% 11
\bibitem[\protect\citeauthoryear{Halbouni et~al.}{2022}]{R8}
\begin{barticle}
\bauthor{\bsnm{Halbouni}, \binits{A.}},
\bauthor{\bsnm{Gunawan}, \binits{T.S.}},
\bauthor{\bsnm{Habaebi}, \binits{M.H.}},
\bauthor{\bsnm{Halbouni}, \binits{M.}},
\bauthor{\bsnm{Kartiwi}, \binits{M.}},
\bauthor{\bsnm{Ahmad}, \binits{R.}}:
\batitle{Cnn-lstm: hybrid deep neural network for network intrusion detection system}.
\bjtitle{IEEE Access}
\bvolume{10},
\bfpage{99837}--\blpage{99849}
(\byear{2022})
\end{barticle}
\endbibitem

%%% 12
\bibitem[\protect\citeauthoryear{Deore and Bhosale}{2022}]{R9}
\begin{barticle}
\bauthor{\bsnm{Deore}, \binits{B.}},
\bauthor{\bsnm{Bhosale}, \binits{S.}}:
\batitle{Hybrid optimization enabled robust cnn-lstm technique for network intrusion detection}.
\bjtitle{IEEE Access}
\bvolume{10},
\bfpage{65611}--\blpage{65622}
(\byear{2022})
\end{barticle}
\endbibitem

%%% 13
\bibitem[\protect\citeauthoryear{Jaggi et~al.}{2015}]{R10}
\begin{bchapter}
\bauthor{\bsnm{Jaggi}, \binits{M.}},
\bauthor{\bsnm{Smith}, \binits{V.}},
\bauthor{\bsnm{Takac}, \binits{M.}},
\bauthor{\bsnm{Terhorst}, \binits{J.}},
\bauthor{\bsnm{Krishnan}, \binits{S.}},
\bauthor{\bsnm{Hoffmann}, \binits{T.}},
\bauthor{\bsnm{Jordan}, \binits{M.I.}}:
\bctitle{Communication-efficient distributed dual coordinate ascent}.
In: \bbtitle{Advances in Neural Information Processing Systems},
vol. \bseriesno{27}
(\byear{2015})
\end{bchapter}
\endbibitem

%%% 14
\bibitem[\protect\citeauthoryear{Kingma and Ba}{2015}]{R11}
\begin{bchapter}
\bauthor{\bsnm{Kingma}, \binits{D.P.}},
\bauthor{\bsnm{Ba}, \binits{J.}}:
\bctitle{Adam: A method for stochastic optimization}.
In: \bbtitle{3rd International Conference for Learning Representations},
\bconflocation{San Diego}
(\byear{2015})
\end{bchapter}
\endbibitem

%%% 15
\bibitem[\protect\citeauthoryear{Recht et~al.}{2011}]{R12}
\begin{bchapter}
\bauthor{\bsnm{Recht}, \binits{B.}},
\bauthor{\bsnm{Re}, \binits{C.}},
\bauthor{\bsnm{Wright}, \binits{S.}},
\bauthor{\bsnm{Niu}, \binits{F.}}:
\bctitle{Hogwild: A lock-free approach to parallelizing stochastic gradient descent}.
In: \bbtitle{Advances in Neural Information Processing Systems (NIPS)},
vol. \bseriesno{24}
(\byear{2011})
\end{bchapter}
\endbibitem

%%% 16
\bibitem[\protect\citeauthoryear{de~Visser et~al.}{2020}]{R13}
\begin{barticle}
\bauthor{\bsnm{Visser}, \binits{E.J.}},
\bauthor{\bsnm{Peeters}, \binits{M.M.}},
\bauthor{\bsnm{Jung}, \binits{M.F.}},
\bauthor{\bsnm{Kohn}, \binits{S.}},
\bauthor{\bsnm{Shaw}, \binits{T.H.}},
\bauthor{\bsnm{Pak}, \binits{R.}},
\bauthor{\bsnm{Neerincx}, \binits{M.A.}}:
\batitle{Towards a theory of longitudinal trust calibration in human–robot teams}.
\bjtitle{International Journal of Social Robotics}
\bvolume{12}(\bissue{2}),
\bfpage{459}--\blpage{478}
(\byear{2020})
\end{barticle}
\endbibitem

%%% 17
\bibitem[\protect\citeauthoryear{Kwon et~al.}{2019}]{R14}
\begin{botherref}
\oauthor{\bsnm{Kwon}, \binits{D.}},
\oauthor{\bsnm{Kim}, \binits{H.}},
\oauthor{\bsnm{Kim}, \binits{J.}},
\oauthor{\bsnm{Suh}, \binits{S.}},
\oauthor{\bsnm{Kim}, \binits{I.}},
\oauthor{\bsnm{Kim}, \binits{K.J.}}:
A survey of deep learning-based network anomaly detection.
Cluster Computing
(2019)
\end{botherref}
\endbibitem

%%% 18
\bibitem[\protect\citeauthoryear{Sridhar et~al.}{2012}]{R15}
\begin{botherref}
\oauthor{\bsnm{Sridhar}, \binits{S.}},
\oauthor{\bsnm{Hahn}, \binits{A.}},
\oauthor{\bsnm{Govindarasu}, \binits{M.}}:
Cyber physical system security for the electric power grid.
Proceedings of the IEEE
\textbf{100}(1)
(2012)
\end{botherref}
\endbibitem

%%% 19
\bibitem[\protect\citeauthoryear{Rajkumar et~al.}{2010}]{R16}
\begin{bchapter}
\bauthor{\bsnm{Rajkumar}, \binits{R.R.}},
\bauthor{\bsnm{Lee}, \binits{I.}},
\bauthor{\bsnm{Sha}, \binits{L.}},
\bauthor{\bsnm{Stankovic}, \binits{J.}}:
\bctitle{Cyber-physical systems: The next computing revolution}.
In: \bbtitle{Proceedings of the 47th Design Automation Conference (DAC)},
\bconflocation{Anaheim, California, USA},
pp. \bfpage{731}--\blpage{736}
(\byear{2010})
\end{bchapter}
\endbibitem

%%% 20
\bibitem[\protect\citeauthoryear{Cardenas et~al.}{2009}]{R17}
\begin{bchapter}
\bauthor{\bsnm{Cardenas}, \binits{A.}},
\bauthor{\bsnm{Amin}, \binits{S.}},
\bauthor{\bsnm{Sinopoli}, \binits{B.}},
\bauthor{\bsnm{Giani}, \binits{A.}},
\bauthor{\bsnm{Perrig}, \binits{A.}},
\bauthor{\bsnm{Sastry}, \binits{S.}}:
\bctitle{Challenges for securing cyber physical systems}.
In: \bbtitle{Workshop on Future Directions in Cyber-physical Systems Security},
\bconflocation{DHS}
(\byear{2009})
\end{bchapter}
\endbibitem

%%% 21
\bibitem[\protect\citeauthoryear{Lee et~al.}{1999}]{R18}
\begin{bchapter}
\bauthor{\bsnm{Lee}, \binits{W.}},
\bauthor{\bsnm{Stolfo}, \binits{S.}},
\bauthor{\bsnm{Mok}, \binits{K.}}:
\bctitle{A data mining framework for building intrusion detection models}.
In: \bbtitle{Proceedings of the 1999 IEEE Symposium on Security and Privacy},
vol. \bseriesno{00},
pp. \bfpage{120}--\blpage{132}
(\byear{1999})
\end{bchapter}
\endbibitem

%%% 22
\bibitem[\protect\citeauthoryear{Buczak and Guven}{2016}]{R19}
\begin{barticle}
\bauthor{\bsnm{Buczak}, \binits{A.L.}},
\bauthor{\bsnm{Guven}, \binits{E.}}:
\batitle{A survey of data mining and machine learning methods for cyber security intrusion detection}.
\bjtitle{IEEE Communications Surveys and Tutorials}
\bvolume{18}(\bissue{2}),
\bfpage{1153}--\blpage{1176}
(\byear{2016})
\end{barticle}
\endbibitem

%%% 23
\bibitem[\protect\citeauthoryear{Amarasinghe et~al.}{2018}]{R20}
\begin{bchapter}
\bauthor{\bsnm{Amarasinghe}, \binits{K.}},
\bauthor{\bsnm{Kenney}, \binits{K.}},
\bauthor{\bsnm{Manic}, \binits{M.}}:
\bctitle{Toward explainable deep neural network based anomaly detection}.
In: \bbtitle{11th International Conference on Human System Interactions (HSI)}
(\byear{2018}).
\doiurl{10.1109/HSI.2018.8430788}
\end{bchapter}
\endbibitem

%%% 24
\bibitem[\protect\citeauthoryear{Samek et~al.}{2017}]{R21}
\begin{barticle}
\bauthor{\bsnm{Samek}, \binits{W.}},
\bauthor{\bsnm{Wiegand}, \binits{T.}},
\bauthor{\bsnm{Müller}, \binits{K.}}:
\batitle{Explainable artificial intelligence: Understanding, visualizing and interpreting deep learning models}.
\bjtitle{ITU Journal: ICT Discoveries}
\bvolume{1}(\bissue{1}),
\bfpage{39}--\blpage{48}
(\byear{2017})
\end{barticle}
\endbibitem

%%% 25
\bibitem[\protect\citeauthoryear{Gunning and Aha}{2017}]{R22}
\begin{barticle}
\bauthor{\bsnm{Gunning}, \binits{D.}},
\bauthor{\bsnm{Aha}, \binits{D.}}:
\batitle{Darpa's explainable artificial intelligence (xai) program}.
\bjtitle{AI Magazine}
\bvolume{40}(\bissue{2}),
\bfpage{44}--\blpage{58}
(\byear{2017})
\doiurl{10.1609/aimag.v40i2.2850}
\end{barticle}
\endbibitem

%%% 26
\bibitem[\protect\citeauthoryear{Parliament and of~the European~Union}{2016}]{R23}
\begin{botherref}
\oauthor{\bsnm{Parliament}},
\oauthor{\bsnm{European~Union}, \binits{C.}}:
General data protection regulation
(2016)
\end{botherref}
\endbibitem

%%% 27
\bibitem[\protect\citeauthoryear{Goodman and Flaxman}{2017}]{R24}
\begin{barticle}
\bauthor{\bsnm{Goodman}, \binits{B.}},
\bauthor{\bsnm{Flaxman}, \binits{S.}}:
\batitle{European union regulations on algorithmic decision-making and ai}.
\bjtitle{AI Magazine}
\bvolume{38}(\bissue{3}),
\bfpage{50}--\blpage{57}
(\byear{2017})
\end{barticle}
\endbibitem

%%% 28
\bibitem[\protect\citeauthoryear{Moustafa et~al.}{2019}]{R25}
\begin{barticle}
\bauthor{\bsnm{Moustafa}, \binits{N.}},
\bauthor{\bsnm{Turnbull}, \binits{B.}},
\bauthor{\bsnm{Choo}, \binits{K.K.R.}}:
\batitle{An ensemble intrusion detection technique based on proposed statistical flow features for protecting network traffic of internet of things}.
\bjtitle{IEEE Internet of Things Journal}
\bvolume{6}(\bissue{3}),
\bfpage{4815}--\blpage{4830}
(\byear{2019})
\end{barticle}
\endbibitem

%%% 29
\bibitem[\protect\citeauthoryear{da~Costa et~al.}{2020}]{R26}
\begin{botherref}
\oauthor{\bsnm{Costa}, \binits{K.A.P.}},
\oauthor{\bsnm{Papa}, \binits{J.P.}},
\oauthor{\bsnm{Passos}, \binits{L.A.}},
\oauthor{\bsnm{Colombo}, \binits{D.}},
\oauthor{\bsnm{Ser}, \binits{J.D.}},
\oauthor{\bsnm{Muhammad}, \binits{K.}},
\oauthor{\bsnm{Albuquerque}, \binits{V.H.C.}}:
A critical literature survey and prospects on tampering and anomaly detection in image data.
Applied Soft Computing
\textbf{97}
(2020)
\end{botherref}
\endbibitem

%%% 30
\bibitem[\protect\citeauthoryear{Kumar and Alqahtani}{2023}]{E4}
\begin{botherref}
\oauthor{\bsnm{Kumar}, \binits{G.}},
\oauthor{\bsnm{Alqahtani}, \binits{H.}}:
Machine learning techniques for intrusion detection systems in sdn-recent advances, challenges and future directions.
CMES-Computer Modeling in Engineering and Sciences
(2023)
\end{botherref}
\endbibitem

%%% 31
\bibitem[\protect\citeauthoryear{Rao and Mane}{2013}]{R27}
\begin{botherref}
\oauthor{\bsnm{Rao}, \binits{D.}},
\oauthor{\bsnm{Mane}, \binits{S.}}:
European union regulations on algorithmic decision-making and AI.
arXiv preprint arXiv:2106.14647
(2013)
\end{botherref}
\endbibitem

%%% 32
\bibitem[\protect\citeauthoryear{Li et~al.}{2018}]{R28}
\begin{barticle}
\bauthor{\bsnm{Li}, \binits{S.}},
\bauthor{\bsnm{Xu}, \binits{L.D.}},
\bauthor{\bsnm{Zhao}, \binits{S.}}:
\batitle{5g internet of things: A survey}.
\bjtitle{Journal of Industrial Information Integration}
\bvolume{10},
\bfpage{19}
(\byear{2018})
\end{barticle}
\endbibitem

%%% 33
\bibitem[\protect\citeauthoryear{Santos et~al.}{2018}]{R29}
\begin{bchapter}
\bauthor{\bsnm{Santos}, \binits{L.}},
\bauthor{\bsnm{Rabadao}, \binits{C.}},
\bauthor{\bsnm{Goncalves}, \binits{R.}}:
\bctitle{Intrusion detection systems in internet of things: A literature review}.
In: \bbtitle{Proceedings of the 13th Iberian Conference on Information Systems and Technologies (CISTI)},
pp. \bfpage{1}--\blpage{7}
(\byear{2018})
\end{bchapter}
\endbibitem

%%% 34
\bibitem[\protect\citeauthoryear{Conti et~al.}{2018}]{R30}
\begin{barticle}
\bauthor{\bsnm{Conti}, \binits{M.}},
\bauthor{\bsnm{Dehghantanha}, \binits{A.}},
\bauthor{\bsnm{Franke}, \binits{K.}},
\bauthor{\bsnm{Watson}, \binits{S.}}:
\batitle{Internet of things security and forensics: Challenges and opportunities}.
\bjtitle{Future Generation Computer Systems}
\bvolume{78},
\bfpage{544}--\blpage{546}
(\byear{2018})
\end{barticle}
\endbibitem

%%% 35
\bibitem[\protect\citeauthoryear{Buczak and Guven}{2016}]{R31}
\begin{barticle}
\bauthor{\bsnm{Buczak}, \binits{A.L.}},
\bauthor{\bsnm{Guven}, \binits{E.}}:
\batitle{A survey of data mining and machine learning methods for cyber security intrusion detection}.
\bjtitle{IEEE Communications Surveys \& Tutorials}
\bvolume{18}(\bissue{2}),
\bfpage{1153}--\blpage{1176}
(\byear{2016})
\end{barticle}
\endbibitem

%%% 36
\bibitem[\protect\citeauthoryear{Arya and Mishra}{2011}]{R32}
\begin{barticle}
\bauthor{\bsnm{Arya}, \binits{I.B.}},
\bauthor{\bsnm{Mishra}, \binits{R.}}:
\batitle{Internet traffic classification: An enhancement in performance using classifiers combination}.
\bjtitle{International Journal of Computer Science and Information Technology}
\bvolume{2}(\bissue{2}),
\bfpage{663}--\blpage{667}
(\byear{2011})
\end{barticle}
\endbibitem

%%% 37
\bibitem[\protect\citeauthoryear{Malik and Khan}{2018}]{R33}
\begin{barticle}
\bauthor{\bsnm{Malik}, \binits{A.J.}},
\bauthor{\bsnm{Khan}, \binits{F.A.}}:
\batitle{A hybrid technique using binary particle swarm optimization and decision tree pruning for network intrusion detection}.
\bjtitle{Cluster Computing}
\bvolume{21}(\bissue{1}),
\bfpage{667}--\blpage{680}
(\byear{2018})
\end{barticle}
\endbibitem

%%% 38
\bibitem[\protect\citeauthoryear{Ahmim et~al.}{2019}]{R34}
\begin{bchapter}
\bauthor{\bsnm{Ahmim}, \binits{A.}},
\bauthor{\bsnm{Maglaras}, \binits{L.}},
\bauthor{\bsnm{Ferrag}, \binits{M.A.}},
\bauthor{\bsnm{Derdour}, \binits{M.}},
\bauthor{\bsnm{Janicke}, \binits{H.}}:
\bctitle{A novel hierarchical intrusion detection system based on decision tree and rules-based models}.
In: \bbtitle{Proceedings of the 15th International Conference on Distributed Computing in Sensor Systems (DCOSS)},
pp. \bfpage{228}--\blpage{233}
(\byear{2019})
\end{bchapter}
\endbibitem

%%% 39
\bibitem[\protect\citeauthoryear{Heba et~al.}{2010}]{R35}
\begin{bchapter}
\bauthor{\bsnm{Heba}, \binits{F.E.}},
\bauthor{\bsnm{Darwish}, \binits{A.}},
\bauthor{\bsnm{Hassanien}, \binits{A.E.}},
\bauthor{\bsnm{Abraham}, \binits{A.}}:
\bctitle{Principle components analysis and support vector machine based intrusion detection system}.
In: \bbtitle{Proceedings of the 10th International Conference on Intelligent Systems Design and Applications},
pp. \bfpage{363}--\blpage{367}
(\byear{2010})
\end{bchapter}
\endbibitem

%%% 40
\bibitem[\protect\citeauthoryear{Kabir et~al.}{2018}]{R36}
\begin{barticle}
\bauthor{\bsnm{Kabir}, \binits{E.}},
\bauthor{\bsnm{Hu}, \binits{J.}},
\bauthor{\bsnm{Wang}, \binits{H.}},
\bauthor{\bsnm{Zhuo}, \binits{G.}}:
\batitle{A novel statistical technique for intrusion detection systems}.
\bjtitle{Future Generation Computer Systems}
\bvolume{79},
\bfpage{303}--\blpage{318}
(\byear{2018})
\end{barticle}
\endbibitem

%%% 41
\bibitem[\protect\citeauthoryear{Agarap}{2018}]{R37}
\begin{bchapter}
\bauthor{\bsnm{Agarap}, \binits{A.F.M.}}:
\bctitle{A neural network architecture combining gated recurrent unit (gru) and support vector machine (svm) for intrusion detection in network traffic data}.
In: \bbtitle{Proceedings of the 10th International Conference on Machine Learning and Computing (ICMLC)},
pp. \bfpage{26}--\blpage{30}
(\byear{2018})
\end{bchapter}
\endbibitem

%%% 42
\bibitem[\protect\citeauthoryear{Serpen and Aghaei}{2018}]{R38}
\begin{barticle}
\bauthor{\bsnm{Serpen}, \binits{G.}},
\bauthor{\bsnm{Aghaei}, \binits{E.}}:
\batitle{Host-based misuse intrusion detection using pca feature extraction and knn classification algorithms}.
\bjtitle{Intelligent Data Analysis}
\bvolume{22}(\bissue{5}),
\bfpage{1101}--\blpage{1114}
(\byear{2018})
\end{barticle}
\endbibitem

%%% 43
\bibitem[\protect\citeauthoryear{Zhang et~al.}{2018}]{R39}
\begin{barticle}
\bauthor{\bsnm{Zhang}, \binits{B.}},
\bauthor{\bsnm{Liu}, \binits{Z.}},
\bauthor{\bsnm{Jia}, \binits{Y.}},
\bauthor{\bsnm{Ren}, \binits{J.}},
\bauthor{\bsnm{Zhao}, \binits{X.}}:
\batitle{Network intrusion detection method based on pca and bayes algorithm}.
\bjtitle{Security and Communication Networks}
\bvolume{2018},
\bfpage{1}--\blpage{11}
(\byear{2018})
\end{barticle}
\endbibitem

%%% 44
\bibitem[\protect\citeauthoryear{Srinivasu and Avadhani}{2012}]{R40}
\begin{barticle}
\bauthor{\bsnm{Srinivasu}, \binits{P.}},
\bauthor{\bsnm{Avadhani}, \binits{P.S.}}:
\batitle{Genetic algorithm based weight extraction algorithm for artificial neural network classifier in intrusion detection}.
\bjtitle{Procedia Engineering}
\bvolume{38},
\bfpage{144}--\blpage{153}
(\byear{2012})
\end{barticle}
\endbibitem

%%% 45
\bibitem[\protect\citeauthoryear{Thaseen and Kumar}{2017}]{R41}
\begin{barticle}
\bauthor{\bsnm{Thaseen}, \binits{I.S.}},
\bauthor{\bsnm{Kumar}, \binits{C.A.}}:
\batitle{Intrusion detection model using fusion of chi-square feature selection and multi-class svm}.
\bjtitle{Journal of King Saud University - Computer and Information Sciences}
\bvolume{29}(\bissue{4}),
\bfpage{462}--\blpage{472}
(\byear{2017})
\end{barticle}
\endbibitem

%%% 46
\bibitem[\protect\citeauthoryear{Thaseen et~al.}{2019}]{R42}
\begin{barticle}
\bauthor{\bsnm{Thaseen}, \binits{I.S.}},
\bauthor{\bsnm{Kumar}, \binits{C.A.}},
\bauthor{\bsnm{Ahmad}, \binits{A.}}:
\batitle{Integrated intrusion detection model using chi-square feature selection and ensemble of classifiers}.
\bjtitle{Arabian Journal of Science and Engineering}
\bvolume{44}(\bissue{4}),
\bfpage{3357}--\blpage{3368}
(\byear{2019})
\end{barticle}
\endbibitem

%%% 47
\bibitem[\protect\citeauthoryear{Alqahtani and Kumar}{2022}]{E1}
\begin{botherref}
\oauthor{\bsnm{Alqahtani}, \binits{H.}},
\oauthor{\bsnm{Kumar}, \binits{G.}}:
A deep learning-based intrusion detection system for in-vehicle networks.
Computers and Electrical Engineering
(2022)
\end{botherref}
\endbibitem

%%% 48
\bibitem[\protect\citeauthoryear{Besharati et~al.}{2019}]{R43}
\begin{barticle}
\bauthor{\bsnm{Besharati}, \binits{E.}},
\bauthor{\bsnm{Naderan}, \binits{M.}},
\bauthor{\bsnm{Namjoo}, \binits{E.}}:
\batitle{Lr-hids: Logistic regression host-based intrusion detection system for cloud environments}.
\bjtitle{Journal of Ambient Intelligence and Humanized Computing}
\bvolume{10}(\bissue{9}),
\bfpage{3669}--\blpage{3692}
(\byear{2019})
\end{barticle}
\endbibitem

%%% 49
\bibitem[\protect\citeauthoryear{Liu et~al.}{2021}]{R44}
\begin{botherref}
\oauthor{\bsnm{Liu}, \binits{M.}},
\oauthor{\bsnm{Xue}, \binits{Z.}},
\oauthor{\bsnm{He}, \binits{X.}},
\oauthor{\bsnm{Chen}, \binits{J.}}:
Scads: A scalable approach using spark in cloud for host-based intrusion detection system with system calls.
arXiv preprint arXiv:2109.11821
(2021)
\end{botherref}
\endbibitem

%%% 50
\bibitem[\protect\citeauthoryear{Park et~al.}{2021}]{R45}
\begin{barticle}
\bauthor{\bsnm{Park}, \binits{D.}},
\bauthor{\bsnm{Kim}, \binits{S.}},
\bauthor{\bsnm{Kwon}, \binits{H.}},
\bauthor{\bsnm{Shin}, \binits{D.}},
\bauthor{\bsnm{Shin}, \binits{D.}}:
\batitle{Host-based intrusion detection model using siamese network}.
\bjtitle{IEEE Access}
\bvolume{9},
\bfpage{76614}--\blpage{76623}
(\byear{2021})
\end{barticle}
\endbibitem

%%% 51
\bibitem[\protect\citeauthoryear{Yu and Bian}{2020}]{R46}
\begin{barticle}
\bauthor{\bsnm{Yu}, \binits{Y.}},
\bauthor{\bsnm{Bian}, \binits{N.}}:
\batitle{An intrusion detection method using few-shot learning}.
\bjtitle{IEEE Access}
\bvolume{8},
\bfpage{49730}--\blpage{49740}
(\byear{2020})
\end{barticle}
\endbibitem

%%% 52
\bibitem[\protect\citeauthoryear{Yang et~al.}{2020}]{R47}
\begin{barticle}
\bauthor{\bsnm{Yang}, \binits{Y.}},
\bauthor{\bsnm{Zheng}, \binits{K.}},
\bauthor{\bsnm{Wu}, \binits{B.}},
\bauthor{\bsnm{Yang}, \binits{Y.}},
\bauthor{\bsnm{Wang}, \binits{X.}}:
\batitle{Network intrusion detection based on supervised adversarial variational auto-encoder with regularization}.
\bjtitle{IEEE Access}
\bvolume{8},
\bfpage{42169}--\blpage{42184}
(\byear{2020})
\end{barticle}
\endbibitem

%%% 53
\bibitem[\protect\citeauthoryear{Clements et~al.}{2021}]{R48}
\begin{bchapter}
\bauthor{\bsnm{Clements}, \binits{J.}},
\bauthor{\bsnm{Yang}, \binits{Y.}},
\bauthor{\bsnm{Sharma}, \binits{A.A.}},
\bauthor{\bsnm{Hu}, \binits{H.}},
\bauthor{\bsnm{Lao}, \binits{Y.}}:
\bctitle{Rallying adversarial techniques against deep learning for network security}.
In: \bbtitle{Proceedings of the IEEE Symposium Series on Computational Intelligence (SSCI)},
pp. \bfpage{01}--\blpage{08}
(\byear{2021})
\end{bchapter}
\endbibitem

%%% 54
\bibitem[\protect\citeauthoryear{Reyes et~al.}{2020}]{R49}
\begin{barticle}
\bauthor{\bsnm{Reyes}, \binits{A.A.}},
\bauthor{\bsnm{Vaca}, \binits{F.D.}},
\bauthor{\bsnm{Aguayo}, \binits{G.A.C.}},
\bauthor{\bsnm{Niyaz}, \binits{Q.}},
\bauthor{\bsnm{Devabhaktuni}, \binits{V.}}:
\batitle{A machine learning based two-stage wi-fi network intrusion detection system}.
\bjtitle{Electronics}
\bvolume{9}(\bissue{10}),
\bfpage{1689}
(\byear{2020})
\end{barticle}
\endbibitem

%%% 55
\bibitem[\protect\citeauthoryear{Kocher and Kumar}{2022}]{E2}
\begin{botherref}
\oauthor{\bsnm{Kocher}, \binits{G.}},
\oauthor{\bsnm{Kumar}, \binits{G.}}:
A hybrid deep learning approach for effective intrusion detection systems using spatial-temporal features.
Adv. Eng. Sci.,
1503--1519
(2022)
\end{botherref}
\endbibitem

%%% 56
\bibitem[\protect\citeauthoryear{Sarhan et~al.}{2021}]{R50}
\begin{botherref}
\oauthor{\bsnm{Sarhan}, \binits{M.}},
\oauthor{\bsnm{Layeghy}, \binits{S.}},
\oauthor{\bsnm{Portmann}, \binits{M.}}:
Evaluating standard feature sets towards increased generalisability and explainability of ml-based network intrusion detection.
arXiv preprint arXiv:2104.07183
(2021)
\end{botherref}
\endbibitem

%%% 57
\bibitem[\protect\citeauthoryear{Mahbooba et~al.}{2021}]{R51}
\begin{barticle}
\bauthor{\bsnm{Mahbooba}, \binits{B.}},
\bauthor{\bsnm{Sahal}, \binits{R.}},
\bauthor{\bsnm{Alosaimi}, \binits{W.}},
\bauthor{\bsnm{Serrano}, \binits{M.}}:
\batitle{Trust in intrusion detection systems: An investigation of performance analysis for machine learning and deep learning models}.
\bjtitle{Complexity}
\bvolume{2021},
\bfpage{1}--\blpage{23}
(\byear{2021})
\end{barticle}
\endbibitem

%%% 58
\bibitem[\protect\citeauthoryear{Li et~al.}{2022}]{R52}
\begin{barticle}
\bauthor{\bsnm{Li}, \binits{S.}},
\bauthor{\bsnm{Zhou}, \binits{Q.}},
\bauthor{\bsnm{Zhou}, \binits{R.}},
\bauthor{\bsnm{Lv}, \binits{Q.}}:
\batitle{Intelligent malware detection based on graph convolutional network}.
\bjtitle{Journal of Supercomputing}
\bvolume{78}(\bissue{3}),
\bfpage{4182}--\blpage{4198}
(\byear{2022})
\end{barticle}
\endbibitem

%%% 59
\bibitem[\protect\citeauthoryear{Kim et~al.}{2017}]{R53}
\begin{bchapter}
\bauthor{\bsnm{Kim}, \binits{J.}},
\bauthor{\bsnm{Shin}, \binits{N.}},
\bauthor{\bsnm{Jo}, \binits{S.Y.}},
\bauthor{\bsnm{Kim}, \binits{S.H.}}:
\bctitle{Method of intrusion detection using deep neural network}.
In: \bbtitle{Proceedings of the IEEE International Conference on Big Data and Smart Computing (BigComp)},
pp. \bfpage{313}--\blpage{316}
(\byear{2017})
\end{bchapter}
\endbibitem

%%% 60
\bibitem[\protect\citeauthoryear{Sölch}{2015}]{R54}
\begin{botherref}
\oauthor{\bsnm{Sölch}, \binits{M.}}:
Detecting anomalies in robot time series data using stochastic recurrent networks.
Master's thesis,
Department of Mathematics, Technische Universität München,
München, Germany
(2015)
\end{botherref}
\endbibitem

%%% 61
\bibitem[\protect\citeauthoryear{Jiang et~al.}{2022}]{R55}
\begin{barticle}
\bauthor{\bsnm{Jiang}, \binits{K.}},
\bauthor{\bsnm{Wang}, \binits{W.}},
\bauthor{\bsnm{Wang}, \binits{A.}},
\bauthor{\bsnm{Wu}, \binits{H.}}:
\batitle{Network intrusion detection combined hybrid sampling with deep hierarchical network}.
\bjtitle{IEEE Access}
\bvolume{8},
\bfpage{32464}--\blpage{32476}
(\byear{2022})
\end{barticle}
\endbibitem

%%% 62
\bibitem[\protect\citeauthoryear{Wang et~al.}{2021}]{R56}
\begin{bchapter}
\bauthor{\bsnm{Wang}, \binits{Y.}},
\bauthor{\bsnm{Wang}, \binits{P.}},
\bauthor{\bsnm{Wang}, \binits{Z.}},
\bauthor{\bsnm{Cao}, \binits{M.}}:
\bctitle{An explainable intrusion detection system}.
In: \bbtitle{Proceedings of the IEEE 23rd International Conference on High Performance Computing and Communications, 7th International Conference on Data Science and Systems, 19th International Conference on Smart City, 7th International Conference on Dependability of Sensor, Cloud, Big Data Systems and Applications (HPCC/DSS/SmartCity/DependSys)},
pp. \bfpage{1657}--\blpage{1662}
(\byear{2021})
\end{bchapter}
\endbibitem

%%% 63
\bibitem[\protect\citeauthoryear{Wali and Khan}{2021}]{R57}
\begin{barticle}
\bauthor{\bsnm{Wali}, \binits{S.}},
\bauthor{\bsnm{Khan}, \binits{I.}}:
\batitle{Explainable ai and random forest based reliable intrusion detection system}.
\bjtitle{TechRxiv}
(\byear{2021})
\doiurl{10.36227/techrxiv.17169080.v1}
\end{barticle}
\endbibitem

%%% 64
\bibitem[\protect\citeauthoryear{Tcydenova et~al.}{2021}]{R58}
\begin{barticle}
\bauthor{\bsnm{Tcydenova}, \binits{E.}},
\bauthor{\bsnm{Kim}, \binits{T.W.}},
\bauthor{\bsnm{Lee}, \binits{C.}},
\bauthor{\bsnm{Park}, \binits{J.H.}}:
\batitle{Detection of adversarial attacks in ai-based intrusion detection systems using explainable ai}.
\bjtitle{Human-Centric Computing and Information Sciences}
\bvolume{11},
\bfpage{1}--\blpage{14}
(\byear{2021})
\end{barticle}
\endbibitem

%%% 65
\bibitem[\protect\citeauthoryear{Shibghatullah}{2023}]{C7}
\begin{botherref}
\oauthor{\bsnm{Shibghatullah}, \binits{A.S..B.}}:
Mitigating developed persistent threats (apts) through machine learning-based intrusion detection systems: A comprehensive analysis.
SHIFRA,
17--25
(2023)
\end{botherref}
\endbibitem

%%% 66
\bibitem[\protect\citeauthoryear{Salman and Alsajri}{2023}]{C8}
\begin{botherref}
\oauthor{\bsnm{Salman}, \binits{H.A.}},
\oauthor{\bsnm{Alsajri}, \binits{A.}}:
The evolution of cybersecurity threats and strategies for effective protection. a review.
SHIFRA,
73--85
(2023)
\end{botherref}
\endbibitem

%%% 67
\bibitem[\protect\citeauthoryear{Zebin et~al.}{2022}]{R59}
\begin{barticle}
\bauthor{\bsnm{Zebin}, \binits{T.}},
\bauthor{\bsnm{Rezvy}, \binits{S.}},
\bauthor{\bsnm{Luo}, \binits{Y.}}:
\batitle{An explainable ai-based intrusion detection system for dns over https (doh) attacks}.
\bjtitle{IEEE Transactions on Information Forensics and Security}
\bvolume{17},
\bfpage{2339}--\blpage{2349}
(\byear{2022})
\end{barticle}
\endbibitem

%%% 68
\bibitem[\protect\citeauthoryear{Andresini et~al.}{2022}]{R60}
\begin{botherref}
\oauthor{\bsnm{Andresini}, \binits{G.}},
\oauthor{\bsnm{Appice}, \binits{A.}},
\oauthor{\bsnm{Caforio}, \binits{F.P.}},
\oauthor{\bsnm{Malerba}, \binits{D.}},
\oauthor{\bsnm{Vessio}, \binits{G.}}:
Roulette: A neural attention multi-output model for explainable network intrusion detection.
Expert Systems with Applications
\textbf{201}
(2022)
\end{botherref}
\endbibitem

%%% 69
\bibitem[\protect\citeauthoryear{Mowla et~al.}{2022}]{R61}
\begin{bchapter}
\bauthor{\bsnm{Mowla}, \binits{N.I.}},
\bauthor{\bsnm{Rosell}, \binits{J.}},
\bauthor{\bsnm{Vahidi}, \binits{A.}}:
\bctitle{Dynamic voting based explainable intrusion detection system for in-vehicle network}.
In: \bbtitle{Proceedings of the 24th International Conference on Advanced Communication Technology (ICACT)},
pp. \bfpage{406}--\blpage{411}
(\byear{2022})
\end{bchapter}
\endbibitem

%%% 70
\bibitem[\protect\citeauthoryear{Mahbooba et~al.}{2021}]{R62}
\begin{barticle}
\bauthor{\bsnm{Mahbooba}, \binits{B.}},
\bauthor{\bsnm{Sahal}, \binits{R.}},
\bauthor{\bsnm{Alosaimi}, \binits{W.}},
\bauthor{\bsnm{Serrano}, \binits{M.}}:
\batitle{Trust in intrusion detection systems: An investigation of performance analysis for machine learning and deep learning models}.
\bjtitle{Complexity}
\bvolume{2021},
\bfpage{1}--\blpage{23}
(\byear{2021})
\end{barticle}
\endbibitem

%%% 71
\bibitem[\protect\citeauthoryear{Radanliev et~al.}{2024a}]{C5}
\begin{botherref}
\oauthor{\bsnm{Radanliev}, \binits{P.}},
\oauthor{\bsnm{Santos}, \binits{O.}},
\oauthor{\bsnm{Brandon-Jones}, \binits{A.}}:
Capability hardware enhanced instructions and artificial intelligence bill of materials in trustworthy artificial intelligence systems: analyzing cybersecurity threats, exploits, and vulnerabilities in new software bills of materials with artificial intelligence.
The Journal of Defense Modeling and Simulation
\textbf{10}
(2024)
\end{botherref}
\endbibitem

%%% 72
\bibitem[\protect\citeauthoryear{Radanliev et~al.}{2024b}]{C6}
\begin{botherref}
\oauthor{\bsnm{Radanliev}, \binits{P.}},
\oauthor{\bsnm{Roure}, \binits{D.D.}},
\oauthor{\bsnm{Maple}, \binits{C.}},
\oauthor{\bsnm{Nurse}, \binits{J.R.C.}},
\oauthor{\bsnm{Nicolescu}, \binits{R.}},
\oauthor{\bsnm{Ani}, \binits{U.}}:
Ai security and cyber risk in iot systems.
Frontiers in Big Data
\textbf{7}
(2024)
\end{botherref}
\endbibitem

%%% 73
\bibitem[\protect\citeauthoryear{Lo et~al.}{2022}]{E3}
\begin{botherref}
\oauthor{\bsnm{Lo}, \binits{W.}},
\oauthor{\bsnm{Alqahtani}, \binits{H.}},
\oauthor{\bsnm{Thakur}, \binits{K.}},
\oauthor{\bsnm{Almadhor}, \binits{A.}},
\oauthor{\bsnm{Chander}, \binits{S.}},
\oauthor{\bsnm{Kumar}, \binits{G.}}:
A hybrid deep learning based intrusion detection system using spatial-temporal representation of in-vehicle network traffic.
Vehicular Communications
\textbf{35}
(2022)
\end{botherref}
\endbibitem

%%% 74
\bibitem[\protect\citeauthoryear{Ribeiro et~al.}{2016}]{R63}
\begin{bchapter}
\bauthor{\bsnm{Ribeiro}, \binits{M.T.}},
\bauthor{\bsnm{Singh}, \binits{S.}},
\bauthor{\bsnm{Guestrin}, \binits{C.}}:
\bctitle{Why should i trust you?: Explaining the predictions of any classifier}.
In: \bbtitle{Proceedings of the 22nd ACM SIGKDD International Conference on Knowledge Discovery and Data Mining},
pp. \bfpage{1135}--\blpage{1144}
(\byear{2016})
\end{bchapter}
\endbibitem

%%% 75
\bibitem[\protect\citeauthoryear{Simonyan et~al.}{2013}]{R64}
\begin{botherref}
\oauthor{\bsnm{Simonyan}, \binits{K.}},
\oauthor{\bsnm{Vedaldi}, \binits{A.}},
\oauthor{\bsnm{Zisserman}, \binits{A.}}:
Deep inside convolutional networks: Visualising image classification models and saliency maps.
arXiv preprint arXiv:1312.6034
(2013)
\end{botherref}
\endbibitem

%%% 76
\bibitem[\protect\citeauthoryear{Iyer et~al.}{2018}]{R65}
\begin{bchapter}
\bauthor{\bsnm{Iyer}, \binits{R.}},
\bauthor{\bsnm{Li}, \binits{Y.}},
\bauthor{\bsnm{Li}, \binits{H.}},
\bauthor{\bsnm{Lewis}, \binits{M.}},
\bauthor{\bsnm{Sundar}, \binits{R.}},
\bauthor{\bsnm{Sycara}, \binits{K.}}:
\bctitle{Transparency and explanation in deep reinforcement learning neural networks}.
In: \bbtitle{Proceedings of the 2018 AAAI/ACM Conference on AI, Ethics, and Society},
pp. \bfpage{144}--\blpage{150}
(\byear{2018})
\end{bchapter}
\endbibitem

%%% 77
\bibitem[\protect\citeauthoryear{Mishra et~al.}{2017}]{R66}
\begin{bchapter}
\bauthor{\bsnm{Mishra}, \binits{S.}},
\bauthor{\bsnm{Sturm}, \binits{B.L.}},
\bauthor{\bsnm{Dixon}, \binits{S.}}:
\bctitle{Local interpretable model-agnostic explanations for music content analysis}.
In: \bbtitle{Proceedings of the International Society for Music Information Retrieval Conference (ISMIR)},
pp. \bfpage{537}--\blpage{543}
(\byear{2017})
\end{bchapter}
\endbibitem

%%% 78
\bibitem[\protect\citeauthoryear{Anjomshoae et~al.}{2019}]{R67}
\begin{bchapter}
\bauthor{\bsnm{Anjomshoae}, \binits{S.}},
\bauthor{\bsnm{Främling}, \binits{K.}},
\bauthor{\bsnm{Najjar}, \binits{A.}}:
\bctitle{Explanations of black-box model predictions by contextual importance and utility}.
In: \bbtitle{International Workshop on Explainable, Transparent Autonomous Agents and Multi-Agent Systems},
pp. \bfpage{95}--\blpage{109}
(\byear{2019})
\end{bchapter}
\endbibitem

%%% 79
\bibitem[\protect\citeauthoryear{Wood et~al.}{2019}]{R68}
\begin{botherref}
\oauthor{\bsnm{Wood}, \binits{T.}},
\oauthor{\bsnm{Kelly}, \binits{C.}},
\oauthor{\bsnm{Roberts}, \binits{M.}},
\oauthor{\bsnm{Walsh}, \binits{B.}}:
An interpretable machine learning model of biological age.
F1000Research
\textbf{8}
(2019)
\end{botherref}
\endbibitem

%%% 80
\bibitem[\protect\citeauthoryear{Azevedo et~al.}{2019}]{R69}
\begin{bchapter}
\bauthor{\bsnm{Azevedo}, \binits{T.}},
\bauthor{\bsnm{Passamonti}, \binits{L.}},
\bauthor{\bsnm{Lió}, \binits{P.}},
\bauthor{\bsnm{Toschi}, \binits{N.}}:
\bctitle{A machine learning tool for interpreting differences in cognition using brain features}.
In: \bbtitle{IFIP International Conference on Artificial Intelligence Applications and Innovations},
pp. \bfpage{475}--\blpage{486}
(\byear{2019})
\end{bchapter}
\endbibitem

%%% 81
\bibitem[\protect\citeauthoryear{Amarasinghe and Manic}{2018}]{R70}
\begin{bchapter}
\bauthor{\bsnm{Amarasinghe}, \binits{K.}},
\bauthor{\bsnm{Manic}, \binits{M.}}:
\bctitle{Improving user trust on deep neural networks based intrusion detection systems}.
In: \bbtitle{Proceedings of the IECON 2018 - 44th Annual Conference of the IEEE Industrial Electronics Society},
pp. \bfpage{3262}--\blpage{3268}
(\byear{2018})
\end{bchapter}
\endbibitem

%%% 82
\bibitem[\protect\citeauthoryear{Bach et~al.}{2015}]{R71}
\begin{botherref}
\oauthor{\bsnm{Bach}, \binits{S.}},
\oauthor{\bsnm{Binder}, \binits{A.}},
\oauthor{\bsnm{Montavon}, \binits{G.}},
\oauthor{\bsnm{Klauschen}, \binits{F.}},
\oauthor{\bsnm{Müller}, \binits{K.R.}},
\oauthor{\bsnm{Samek}, \binits{W.}}:
On pixel-wise explanations for non-linear classifier decisions by layer-wise relevance propagation.
PloS One
\textbf{10}(7)
(2015)
\end{botherref}
\endbibitem

%%% 83
\bibitem[\protect\citeauthoryear{Marino et~al.}{2018}]{R72}
\begin{bchapter}
\bauthor{\bsnm{Marino}, \binits{D.L.}},
\bauthor{\bsnm{Wickramasinghe}, \binits{C.S.}},
\bauthor{\bsnm{Manic}, \binits{M.}}:
\bctitle{An adversarial approach for explainable ai in intrusion detection systems}.
In: \bbtitle{IECON 2018-44th Annual Conference of the IEEE Industrial Electronics Society},
pp. \bfpage{3237}--\blpage{3243}
(\byear{2018})
\end{bchapter}
\endbibitem

%%% 84
\bibitem[\protect\citeauthoryear{Li et~al.}{2019}]{R73}
\begin{bchapter}
\bauthor{\bsnm{Li}, \binits{H.}},
\bauthor{\bsnm{Wei}, \binits{F.}},
\bauthor{\bsnm{Hu}, \binits{H.}}:
\bctitle{Enabling dynamic network access control with anomaly-based ids and sdn}.
In: \bbtitle{Proceedings of the ACM International Workshop on Security in Software Defined Networks and Network Function Virtualization},
pp. \bfpage{13}--\blpage{16}
(\byear{2019})
\end{bchapter}
\endbibitem

%%% 85
\bibitem[\protect\citeauthoryear{Guo et~al.}{2018}]{R74}
\begin{bchapter}
\bauthor{\bsnm{Guo}, \binits{W.}},
\bauthor{\bsnm{Mu}, \binits{D.}},
\bauthor{\bsnm{Xu}, \binits{J.}},
\bauthor{\bsnm{Su}, \binits{P.}},
\bauthor{\bsnm{Wang}, \binits{G.}},
\bauthor{\bsnm{Xing}, \binits{X.}}:
\bctitle{Lemna: Explaining deep learning based security applications}.
In: \bbtitle{Proceedings of the 2018 ACM SIGSAC Conference on Computer and Communications Security},
pp. \bfpage{364}--\blpage{379}
(\byear{2018})
\end{bchapter}
\endbibitem

%%% 86
\bibitem[\protect\citeauthoryear{Du et~al.}{2020}]{R75}
\begin{barticle}
\bauthor{\bsnm{Du}, \binits{M.}},
\bauthor{\bsnm{Liu}, \binits{N.}},
\bauthor{\bsnm{Hu}, \binits{X.}}:
\batitle{Techniques for interpretable machine learning}.
\bjtitle{Communications of the ACM}
\bvolume{63}(\bissue{1}),
\bfpage{68}--\blpage{77}
(\byear{2020})
\end{barticle}
\endbibitem

%%% 87
\bibitem[\protect\citeauthoryear{Sovrano and Vitali}{2021}]{R76}
\begin{botherref}
\oauthor{\bsnm{Sovrano}, \binits{F.}},
\oauthor{\bsnm{Vitali}, \binits{F.}}:
An objective metric for explainable ai: How and why to estimate the degree of explainability.
arXiv preprint arXiv:2109.05327
(2021)
\end{botherref}
\endbibitem

%%% 88
\bibitem[\protect\citeauthoryear{Sovrano et~al.}{2020}]{R77}
\begin{bchapter}
\bauthor{\bsnm{Sovrano}, \binits{F.}},
\bauthor{\bsnm{Vitali}, \binits{F.}},
\bauthor{\bsnm{Palmirani}, \binits{M.}}:
\bctitle{Modelling gdpr-compliant explanations for trustworthy ai}.
In: \bbtitle{Electronic Government and the Information Systems Perspective - 9th International Conference, EGOVIS 2020, Bratislava, Slovakia, September 14-17, 2020, Proceedings},
pp. \bfpage{219}--\blpage{233}
(\byear{2020})
\end{bchapter}
\endbibitem

\end{thebibliography}
%% if required, the content of .bbl file can be included here once bbl is generated
%%\input sn-article.bbl

\end{document}